%% file: main.tex
\documentclass[ijoc,nonblindrev]{informs3} 

\OneAndAHalfSpacedXII 

\usepackage{bm}
\usepackage{caption}
\usepackage{subcaption}
\usepackage{rotating}
\usepackage{mathtools}
\usepackage{pgfplots}
\usepackage{algorithm}
\usepackage{algpseudocode}
\usepackage{booktabs}
\usepackage{cases}
\usepackage{enumerate}
\usepackage{hyperref}
\usepackage[normalem]{ulem}

\pgfplotsset{compat=1.17}

\usepackage{xcolor,colortbl}
\usepackage{booktabs}
\usepackage{tikz}
\usetikzlibrary{arrows}
\usetikzlibrary{shapes}
\definecolor{tempcolor}{RGB}{244,244,244}
\usetikzlibrary{positioning}
\tikzset{
    split/.style = {shape=rectangle,
                     draw, align=center,
                     fill=tempcolor},
    clust/.style = {align=center,
                    draw,
                    rectangle}
                     }

\tikzstyle{process} = [rectangle, minimum width=3cm, minimum height=1cm, text centered, draw=black, fill=orange!30]
\tikzstyle{arrow} = [thick,->,>=stealth]

\renewcommand{\algorithmicrequire}{\textbf{Input:}}
\renewcommand{\algorithmicensure}{\textbf{Output:}}

\algblockdefx{FORP}{ENDFORP}[1]%
  {\textbf{for }#1 \textbf{do in parallel}}%
  {\textbf{end for}}
  
\usepackage{natbib}
 \bibpunct[, ]{(}{)}{,}{a}{}{,}%
 \def\bibfont{\small}%
\TheoremsNumberedThrough     

\EquationsNumberedThrough    

\usepackage{multirow}
\usepackage[hang,flushmargin]{footmisc}
\usepackage{ulem}
\newcommand{\minimize}{\mbox{minimize\hspace{4mm} }}
\newcommand{\maximize}{\mbox{maximize\hspace{4mm} }}
\newcommand{\subto}{\mbox{subject to\hspace{4mm}}}

\newcommand\numberthis{\addtocounter{equation}{1}\tag{\theequation}}

\usepackage[capitalize,noabbrev]{cleveref}

\begin{document}


\RUNAUTHOR{Maragno et al.}

\RUNTITLE{Finding Regions of Counterfactual Explanations via Robust Optimization}

\TITLE{Finding Regions of Counterfactual Explanations via Robust Optimization}

\ARTICLEAUTHORS{%
\AUTHOR{Donato Maragno, Jannis Kurtz, Tabea E. R\"ober, Rob Goedhart, {\c{S}}. {\.{I}}lker Birbil, Dick den Hertog}
\AFF{Amsterdam Business School, University of Amsterdam, 1018TV Amsterdam, Netherlands\\ \EMAIL{d.maragno@uva.nl j.kurtz@uva.nl t.e.rober@uva.nl r.goedhart2@uva.nl
s.i.birbil@uva.nl d.denhertog@uva.nl}}

} 

\ABSTRACT{%
Counterfactual explanations play an important role in detecting bias and improving  the explainability of data-driven classification models. A counterfactual explanation (CE) is a minimal perturbed data point for which the decision of the model changes. Most of the existing methods can only provide one CE, which may not be achievable for the user. In this work we derive an iterative method to calculate robust CEs, \textit{i.e.}, CEs that remain valid even after the features are slightly perturbed. To this end, our method provides a whole region of CEs allowing the user to choose a suitable recourse to obtain a desired outcome. We use algorithmic ideas from robust optimization and prove convergence results for the most common machine learning methods including decision trees, tree ensembles, and neural networks. Our experiments show that our method can efficiently generate globally optimal robust CEs for a variety of common data sets and classification models.
}%



\maketitle

\section{Introduction}
Counterfactual explanations, also known as algorithmic recourse, are becoming increasingly popular as a way to explain the decisions made by black-box machine learning (ML) models. Given a factual instance for which we want to derive an explanation, we search for a counterfactual feature combination describing the minimum change in the feature space that will lead to a flipped model prediction. For example, for a person with a rejected loan application, the counterfactual explanation (CE) could be ``if the \textit{annual salary} would increase to 50,000\$, then the \textit{loan application} would be approved.'' This method enables a form of user agency and is therefore particularly attractive in consequential decision making, where the user is directly and indirectly impacted by the outcome of the ML model. 

The first optimization-based approach to generate CEs has been proposed by \citet{Wachter.2018}. Given a trained classifier $h: \mathcal X \to [0,1]$ and a \textit{factual instance} $\hat{\bm{x}}\in \mathcal X$, the aim is to find a \textit{counterfactual} $\bm{\tilde{x}}\in \mathcal X$ that has the shortest distance to $\hat{\bm{x}}$, and has the opposite target. The problem to obtain $\bm{\tilde{x}}$ can be formulated as
\begin{align}
\label{eqn:wachtermodel}
    \underset{{\bm{x}} \in \mathcal{X}\hspace{4mm}}{~\minimize} & d(\hat{\bm{x}}, \bm{x}) \\
    \subto \ & h(\bm{x}) \ge \tau,
\end{align}
where $d(\cdot,\cdot)$ is a distance function, often chosen to be the $\ell_1$-norm or the $\ell_2$-norm, and $\tau\in [0,1]$ is a given threshold parameter for the classification decision. 

Others have built on this work and proposed approaches that generate CEs with increased practical value, primarily by adding constraints to ensure actionability of the proposed changes and generating CEs that are close to the data manifold \citep{Ustun.2019, Russell.2019, Mahajan.2019, Mothilal.2020, maragno2022counterfactual}. 
Nonetheless, the user agency provided by these methods remains theoretical: the generated CEs are exact point solutions that may remain difficult, if not impossible, to implement in practice. A minimal change to the proposed CE could fail to flip the model's prediction, especially since the CEs are close to the decision boundary due to minimizing the distance between $\hat{x}$ and $\tilde{x}$. As a solution, prior work suggests generating several CEs to increase the likelihood of generating at least one attainable solution. Approaches generating several CEs typically require solving the optimization problem multiple times \citep{Russell.2019, Mothilal.2020, Kanamori.2020jn, Karimi.2019fy}, which might heavily affect the optimization time when the number of explanations is large. \citet{maragno2022counterfactual} suggest using incumbent solutions, however, this does not allow to control the quality of sub-optimal solutions. 
On top of that, the added practical value may be unconvincing: each of the CEs is still sensitive to arbitrarily small changes in the actions implemented by the user \citep{Dominguez.2022, pawelczyk2022probabilistically, VIRGOLIN2023103840}.

This problem has been acknowledged in prior work and falls under the discussion of robustness in CEs. In the literature, the concept of robustness in CEs has different meanings: (1) robustness to input perturbations \citep{Slack_neurips21, Artelt.2021} and generating explanations for a group of individuals \cite{carrizosa2021generating}, (2) robustness to model changes \citep{rawal2020algorithmic, Forel.2022, Upadhyay_neurips21, Ferrario.2022, Black_2021, pmlr-v162-dutta22a, bui2022counterfactual}, (3) robustness to hyperparameter selection \citep{Dandle_multiobjective_2020}, and (4) robustness to recourse \citep{pawelczyk2022probabilistically, Dominguez.2022, VIRGOLIN2023103840}. The latter perspective, albeit very user-centered, has so far received only a little attention. Our work focuses on the latter definition of robustness in CEs, specifically, the idea of robustness to recourse. This means that a counterfactual solution should remain valid even if small changes are made to the implemented recourse action. In other words, we aim to define regions of counterfactual solutions that allow the user to choose any point within that region to flip the model prediction. This extends the idea of offering several explanations for the user to choose from, such that not only the suggested point is a counterfactual, but every point in the defined region. Returning to the example we used above, a final explanation may be ``if the \textit{annual salary} would increase to anywhere between 50,000\$ and 54,200\$, then the \textit{loan application} would be approved.'' While existing research has tackled this problem, their solutions are not comprehensive and have room for further improvements. In the remainder of this section, we will explore the related prior work and present our own contributions to this field.

\citet{pawelczyk2022probabilistically} introduce the notion of recourse invalidation rate, which amounts to the proportion of recourse that does not lead to the desired model prediction, \textit{i.e.}, that is invalid. They model the noise around a counterfactual data point with a Gaussian distribution and suggest an approach that ensures the invalidation rate within a specified neighborhood around the counterfactual data point to be no larger than a target recourse invalidation rate. However, their work provides a heuristic solution using a gradient-based approach, which makes it not directly applicable to decision tree models. Additionally, it only provides a probabilistic robustness guarantee. \citet{Dominguez.2022} introduce an approach where the optimal solution is surrounded by an uncertainty set such that every point in the set is a feasible solution. They also model causality between (perturbed) features to obtain a more informative neighborhood of similar points. Given a structural causal model (SCM), they model such perturbations as additive interventions on the factual instance features. The authors design an iterative approach that works only for differentiable classifiers and does not guarantee that the generated recourse actions are adversarially robust.
\citet{VIRGOLIN2023103840} incorporate the possibility of additional intervention to contrast perturbations in their search for CEs. They make a distinction between the features that could be changed and those that should be kept as they are, and introduce the concept of C-setbacks; a subset of perturbations in changeable features that work against the user. Rather than seeking CEs that are not invalidated by C-setbacks, they seek CEs for which the additional intervention cost to overcome the setback is minimal. Perturbations to features that should be kept as they are according to a CE are orthogonal to the direction of the counterfactual, and \citet{VIRGOLIN2023103840} approximate a robustness-score for such features. A drawback of this method is that it is only applicable in situations where additional intervention is possible, and not in situations where (\textit{e.g.}, due to time limitations) only a single recourse is possible.

Our work addresses robustness to recourse by utilizing a robust optimization approach to generate regions of CEs. For a given factual instance, our method generates a CE that is robust to small perturbations. 
This gives the user more flexibility in implementing the recourse and reduces the risk of invalidating it. In this work we consider numerical features, ensuring that small perturbations do not affect the recourse validity, while categorical features are treated as immutable based on user preferences. Additionally, the generated CEs are optimal in terms of their objective distance to the factual instance. The proposed algorithm is proven to converge, ensuring that the optimal solution is reached. This is different from prior work that provides only heuristic algorithms which are not provably able to find the optimal (i.e., closest) counterfactual point with a certain robustness guarantee \citep[\textit{e.g.,}][]{pawelczyk2022probabilistically, Dominguez.2022}. Unlike prior research in this area, our approach is able to provide deterministic robustness guarantees for the CEs generated. Furthermore, our method does not require differentiability of the underlying ML model and is applicable to the tree-based models, which, to the best of our knowledge, has not been done before.

In summary, we make the following contributions:
\begin{itemize}
\item We propose an iterative algorithm that effectively finds global optimal robust CEs for trained decision trees, ensembles of trees, and neural networks. 
\item We prove the convergence of the algorithm for the considered trained models.
\item We demonstrate the power of our algorithm on several datasets and different ML models. We empirically evaluate its convergence performance and compare the robustness as well as the validity of the generated CEs with the prior work in the literature.
\item We release an open-source software called RCE to make the proposed algorithm easily accessible to practitioners. Our software is available in a dedicated \href{https://github.com/donato-maragno/robust-CE}{repository}\footnote{https://github.com/donato-maragno/robust-CE} through which all our results can be reproduced.
\end{itemize}

\section{Robust Counterfactual Explanations}
\label{sec:rec}
\begin{figure}
    \centering
    \includegraphics[scale=0.3]{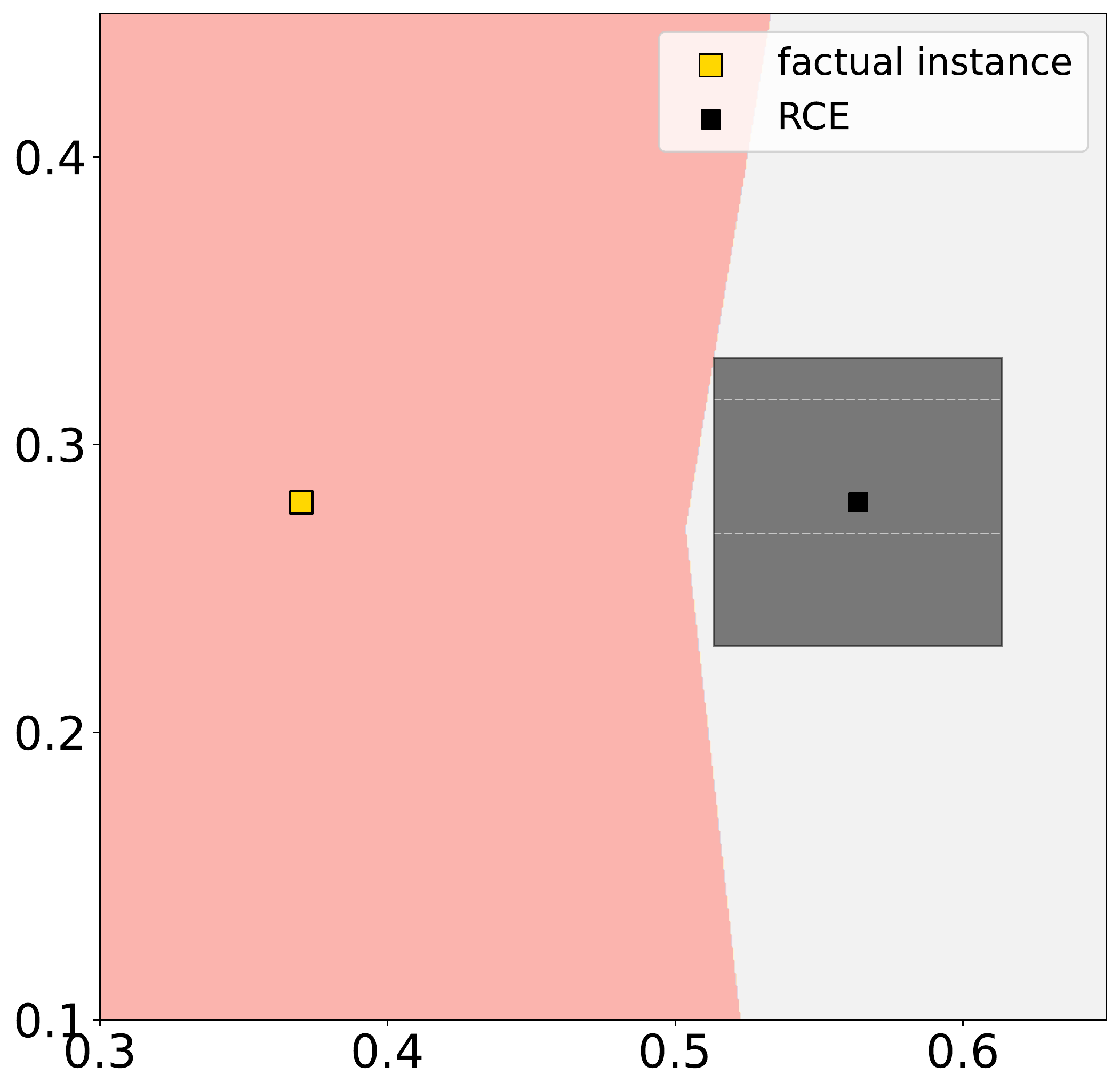}
    \caption{Robust CE for a neural network and using a box uncertainty set. All points in the red region are classified as $-1$, all points in the white region as $+1$.}
    \label{fig:robustCE_NN}
\end{figure}

We consider binary classification problems, \textit{i.e.}, we have a trained classifier $h: \mathcal X\to [0,1]$ that assigns a value between zero and one to each data point in the data space $\mathcal X \subseteq \mathbb R^n$. A point $\bm{x}\in \mathcal X$ is then predicted to correspond to class $+1$, if $h(\bm{x})\ge \tau$ and to class $-1$, otherwise. Here $\tau\in [0,1]$ is a given threshold parameter which is often chosen to be $\tau=0.5$. Given a factual instance $\bm{\hat x}\in \mathcal X$ which is predicted to be in class $-1$, \textit{i.e.}, $h(\bm{\hat x})<\tau$, the robust CE problem is defined as
\begin{align}
    \underset{{\bm{x}\in \mathcal X}\hspace{4mm}}{~\minimize} \ & d(\bm{x}, \hat{\bm{x}}) & \label{eqn:RCE0}\\
    \subto \ & h(\bm{x} + \bm{s}) \ge \tau, 
    ~~~ \forall \bm{s} \in \mathcal{S},\label{eqn:RCE1}
\end{align}
where the  $d(\bm{x}, \hat{\bm{x}})$ represents a distance function, \textit{e.g.}, induced by the $\ell_1$-, $\ell_2$- or $\ell_{\infty}$-norm, and $\mathcal{S}\subset\mathbb R^n$ is a given uncertainty set. The idea of the problem is to find a point that is as close as possible to the factual instance $\bm{\hat x}$ such that for all perturbations $\bm{s}\in\mathcal S$, the corresponding point $\bm{x}+\bm{s}$ is classified as $+1$ which is enforced by constraints (\ref{eqn:RCE1}); see Figure \ref{fig:robustCE_NN}. This results in a large set of counterfactual explanations.

We consider uncertainty sets of the type
\begin{align}
    \mathcal{S} = \{\bm{s}\in\mathbb{R}^n ~~ | ~~ ||\bm{s}|| \leq \rho \},\label{eqn:RCE3}
\end{align}
where $\| \cdot \|$ is a given norm. Popular choices are the $\ell_\infty$-norm, resulting in a box with upper and lower bounds on features, or the $\ell_2$-norm, resulting in a circular uncertainty set. We refer to \citet{BenTal.2009} for a discussion of uncertainty sets. From the user perspective, choosing the $\ell_\infty$-norm has a practical advantage since the region $\mathcal S$ is a box, \textit{i.e.}, we obtain an interval for each attribute of $\bm{\hat x}$. Each attribute can be changed in its corresponding interval independently, resulting in a counterfactual explanation. Hence, the user can easily detect if there exists a CE in the region that can be practically reached. The choice of the robustness budget $\rho$ greatly depends on the specific domain. Larger values of $\rho$ are associated with CEs that are more robust but can have a larger distance to the factual instance. However, our approach outlined in the subsequent sections is designed to minimize the proximity to the factual instance for a given robustness parameter $\rho$. If the perturbation applied to the CE adheres to a known distribution, it becomes feasible to determine $\rho$ in a way that offers a probabilistic guarantee of CE robustness. We refer to Appendix~\ref{app:choice_of_rho} for a comprehensive guide on how to determine the appropriate value for $\rho$.

\subsection{Comparison Against Model-Robustness}\label{sec:comparison_model_rob}
There are several works that study the robustness of counterfactual explanations regarding changes in the parameters of the trained machine learning model; see e.g. \citep{rawal2020algorithmic, Forel.2022, Upadhyay_neurips21, Ferrario.2022, Black_2021, pmlr-v162-dutta22a, bui2022counterfactual}. As a motivation to study this concept the re-training of ML models is often mentioned, where after re-training a model the model-parameters can change.

Consider a classifier $h_{\hat \omega}$ where $\hat \omega$ is the vector of model parameters that was determined during the training process. In the case of a neural network, this vector contains all weights of the neural network, or in the case of a linear classifier, $\omega$ contains all weight parameters of the linear hyperplane separating the two classes.

Translated into the robust optimization setting we study in this work, a model-robust CE is a point, which remains a CE for all parameter values $\omega\in \Omega$, where $\Omega=\left\{ \omega: \| \omega - \hat \omega\|\le \rho_{mod}\right\}$ is a given uncertainty set which contains all possible model parameters which have distance at most $\rho_{mod}$ to the original weights of the classifier. In other words, if $x^{CE}$ is a counterfactual point for the original classifier, \textit{i.e.}, $h_{\hat \omega}(x^{CE})\ge \tau$ then for each classifier $h_\omega$ where $\omega\in\Omega$ this must also be the case, \textit{i.e.}, $h_\omega (x^{CE})\ge \tau$. Note that defining model-robustness for decision tree models is much more elaborate since a decision tree is not only defined by the parameters of its split-hyperplanes but also by the tree structure which can change after re-training the model. So, defining $\Omega$ is not as straightforward as it is for linear models.

However, a natural question arises: Are the two concepts, model-robustness and recourse-robustness, equivalent? In this case concepts from both fields could profit from each other. Unfortunately, this is not the case, which we show in the following example.

Consider the linear classifier given by the hyperplane $0x_1 + x_2 - 1 = 0$, \textit{i.e.}, every point $x\in\mathbb R^2$ where $x_2-1\ge 0$ is classified as $1$ and all others as $0$. Then, for every $\rho,\rho_{mod}>0$ there exists a recourse-robust CE with radius $\rho$ which is not model-robust with radius $\rho_{mod}$.

\begin{figure}
    \centering
    \includegraphics[scale=0.4]{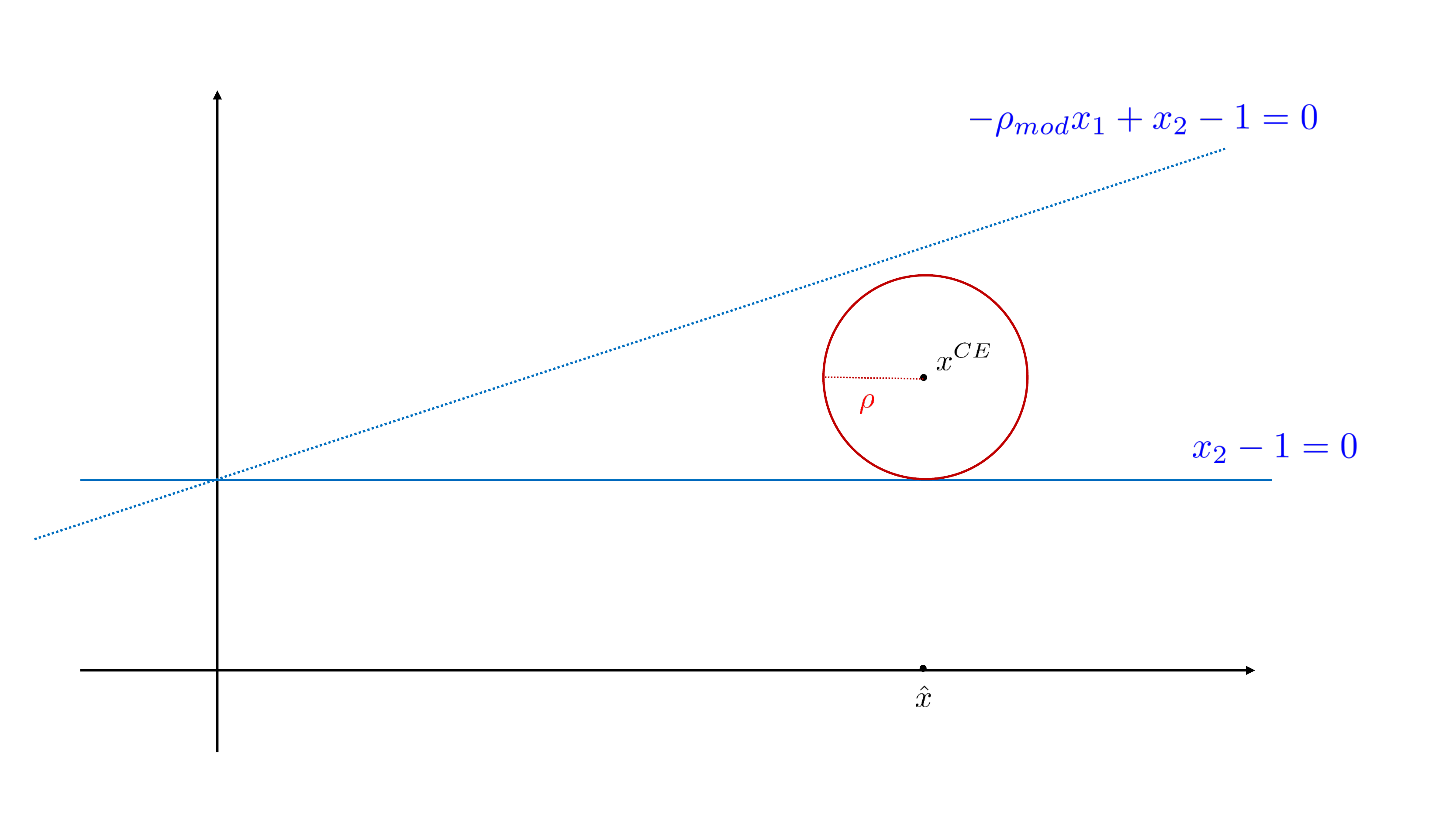}
    \caption{Example in Section \ref{sec:comparison_model_rob}.}
    \label{fig:example_comparison_model_rob}
\end{figure}

The construction works as follows: Let $\rho,\rho_{mod}>0$ and define the factual instance $\hat x=(\frac{\rho}{2\rho_{mod}}, 0)$. Then the closest recourse-robust CE for radius $\rho$ is $x^{CE}=(\frac{\rho}{2\rho_{mod}}, 1+\rho)$ for all relevant norms used in \eqref{eqn:RCE3}; see Figure \ref{fig:example_comparison_model_rob}. Now consider the $\rho_{mod}$-perturbed hyperplane $-\rho_{mod}x_1 + x_2 - 1 = 0$. For $x^{CE}$, it holds
\[
-\rho_{mod}x_1 + x_2 - 1 = -\rho_{mod}\frac{\rho}{2\rho_{mod}} + 1 + \rho - 1 = -\frac{1}{2}\rho<0 .
\]
Hence $x^{CE}$ is not a counterfactual point for the perturbed model.

While the latter example shows that the equivalence of both robustness types does not hold, there can be special cases of models where both concepts are related. However, we place this interesting analysis on our future research agenda.

\subsection{Algorithm}
We note that the model in \eqref{eqn:RCE0}-\eqref{eqn:RCE1} has infinitely many constraints. One approach often used in robust optimization is to rewrite constraints \eqref{eqn:RCE1} as
\[
\min_{\bm{s}\in\mathcal S} h(\bm{x} + \bm{s}) \ge \tau,
\]
and dualize the optimization problem on the left hand side. This leads to a problem with a finite number of constraints. Unfortunately, strong duality is required to perform this reformulation, which does not hold for most classifiers $h$ involving non-convexity or integer variables\footnote{See Appendix~\ref{app:linear_model} for the well-known dual approach applied to linear models.}. In the latter case, we can use an alternative method popular in robust optimization where the constraints are generated iteratively. This iterative approach to solve problem \eqref{eqn:RCE0}-\eqref{eqn:RCE1} is known as the \textit{adversarial approach}. The approach was intensively used for robust optimization problems; see \citet{bienstock2008computing,mutapcic2009cutting}. In \citet{bertsimas2016reformulation} the adversarial approach was compared to the classical robust reformulation.

The idea of the approach is to consider a relaxed version of the model, where only a finite subset of scenarios $\mathcal{Z}\subset \mathcal S$ is considered:
\begin{align}\label{eqn:master_problem}\tag{MP}
    &\underset{{\bm{x}\in \mathcal X}\hspace{4mm}}{\minimize} \ d(\bm{x}, \hat{\bm{x}}) \\
    &\subto \  h(\bm{x} + \bm{s}) \ge \tau, ~~~ \forall \bm{s} \in \mathcal{Z}. \numberthis  \label{eqn:master_problem_constr}
\end{align}
This problem is called the \textit{master problem} (\ref{eqn:master_problem}), and it only has a finite number of constraints. Note that the optimal value of (\ref{eqn:master_problem}) is a lower bound of the optimal value of \eqref{eqn:RCE0}-\eqref{eqn:RCE1}. However, an optimal solution $\bm{x}^*$ of (\ref{eqn:master_problem}) is not necessarily feasible for the original problem, since there may exist a scenario in $\mathcal S$ that is not contained in $\mathcal Z$ for which the solution is not feasible. More precisely, it may be that there exists an $\bm{s}\in\mathcal S$ such that $h(\bm{x}^*+\bm{s})<\tau$, and hence, $\bm{x}^*$ is not a robust counterfactual. 
\begin{figure}
    \centering
    \includegraphics[scale=0.3]{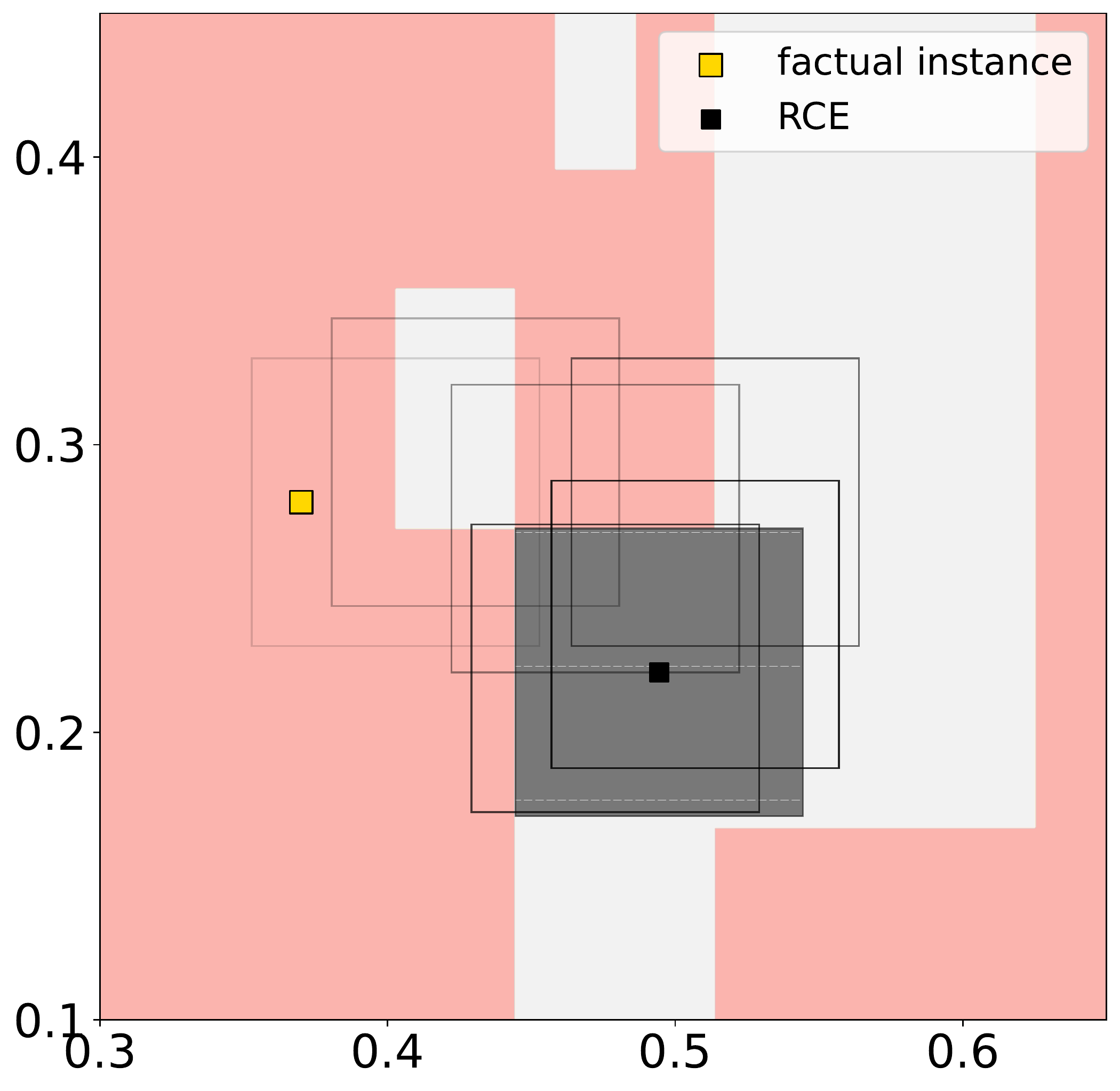}
    \caption{Iterations of Algorithm~\ref{alg:adv_algo} to find the optimal robust CE for a decision tree. For each (\ref{eqn:master_problem}) solution, we show the uncertainty box around it. As long as the box overlaps with the red region, a new scenario can be found and the solution moves in the next iteration. }
    \label{fig:iterationsBox}
\end{figure}
In this case, we want to find such a scenario $\bm{s}^*$ that makes solution $\bm{x}^*$ infeasible. This can be done by solving the following, so called, \textit{adversarial problem} (\ref{eqn:adversarial_problem}):
\begin{equation}\label{eqn:adversarial_problem}\tag{AP}
\begin{aligned}
    &\underset{{\bm{s}\in \mathcal S}}{\max} \ \tau - h(\bm{x}^* + \bm{s}).
\end{aligned}
\end{equation}
The idea is to find a scenario $\bm{s}^*\in \mathcal S$ such that the prediction of classifier $h$ for point $\bm{x}^*+\bm{s}^*$ is $-1$, \textit{i.e.}, $\tau - h(\bm{x}^*+\bm{s}^*)>0$. If we can find such a scenario and add it to the set $\mathcal Z$ in the MP, then $\bm{x}^*$ cannot be feasible anymore for (\ref{eqn:master_problem}). To find the scenario with the largest impact, we maximize the constraint violation in the objective function in (\ref{eqn:adversarial_problem}). If the optimal value of (\ref{eqn:adversarial_problem}) is positive then $\bm{x}^*+\bm{s}^*$ is classified as $-1$, and the optimal solution $\bm{s}^*$ is added to $\mathcal Z$, and we calculate a solution $\bm{x}^*$ of the updated (\ref{eqn:master_problem}). We iterate until no violating scenario can be found, that is, until the optimal value of (\ref{eqn:adversarial_problem}) is smaller or equal to zero. Note that in this case $h(\bm{x}^*+\bm{s})\ge \tau$ holds for all $\bm{s}\in \mathcal S$, which means that $\bm{x}^*$ is a robust counterfactual. Algorithm \ref{alg:adv_algo} shows the steps of our  approach, and Figure \ref{fig:iterationsBox} shows its iterative behaviour. Each time a new scenario $\bm{s}$ is found by solving the AP, it is added to the uncertainty set $\mathcal{Z}$, and the new solution $\bm{x}^*$ moves to be feasible also for the new scenario. This is repeated until no scenario can be found anymore, \textit{i.e.}, until the full box lies in the correct region. Note that instead of checking for a positive optimal value of (\ref{eqn:adversarial_problem}), we use an accuracy parameter $\varepsilon>0$ in Algorithm \ref{alg:adv_algo}. In this case, we can guarantee the convergence of our algorithm using the following result.

\begin{algorithm}[tb]
   \caption{Adversarial Algorithm}
   \label{alg:adv_algo}
\begin{algorithmic}
   \State {\bfseries Input:} $\mathcal{S}$, $\bm{\hat{x}}$, $\varepsilon>0$
   \State $\mathcal Z=\left\{ \bm{0}\right\}$
   \Repeat
   \State $\bm{x}^* \gets$ Solve \eqref{eqn:master_problem} with $\mathcal{Z}, \bm{\hat{x}}$ \label{step:MP}
   \State $\bm{s}^*, \text{opt} \gets$ Solve \eqref{eqn:adversarial_problem} with $\bm{x}^*, \mathcal S$
   \State $\mathcal{Z}\leftarrow \mathcal{Z}\cup \left\{\bm{s}^*\right\}$
   \Until{$\text{opt}\le\varepsilon$}
   \State {\bfseries Return:} $x^*$
\end{algorithmic}
\end{algorithm}

\begin{theorem}[\citeauthor{mutapcic2009cutting}, 2009]
\label{thm:cnvgthm}
If $\mathcal{X}$ is bounded and if $h$ is a Lipschitz continuous function, \textit{i.e.}, there exists an $L>0$ such that
\[
|h(\bm{x_1})-h(\bm{x_2})|\le L \| \bm{x_1}-\bm{x_2}\|
\]
for all $\bm{x_1},\bm{x_2}\in \mathcal{X}$. Then, for any tolerance parameter value $\epsilon > 0$, Algorithm \ref{alg:adv_algo} terminates after a finite number of steps with a solution $\bm{x}^*$ such that $$h(\bm{x}^*+\bm{s})\ge \tau -\varepsilon$$ for all $\bm{s}\in\mathcal S$.
\end{theorem}

\begin{figure}
    \centering
    \includegraphics[scale=0.3]{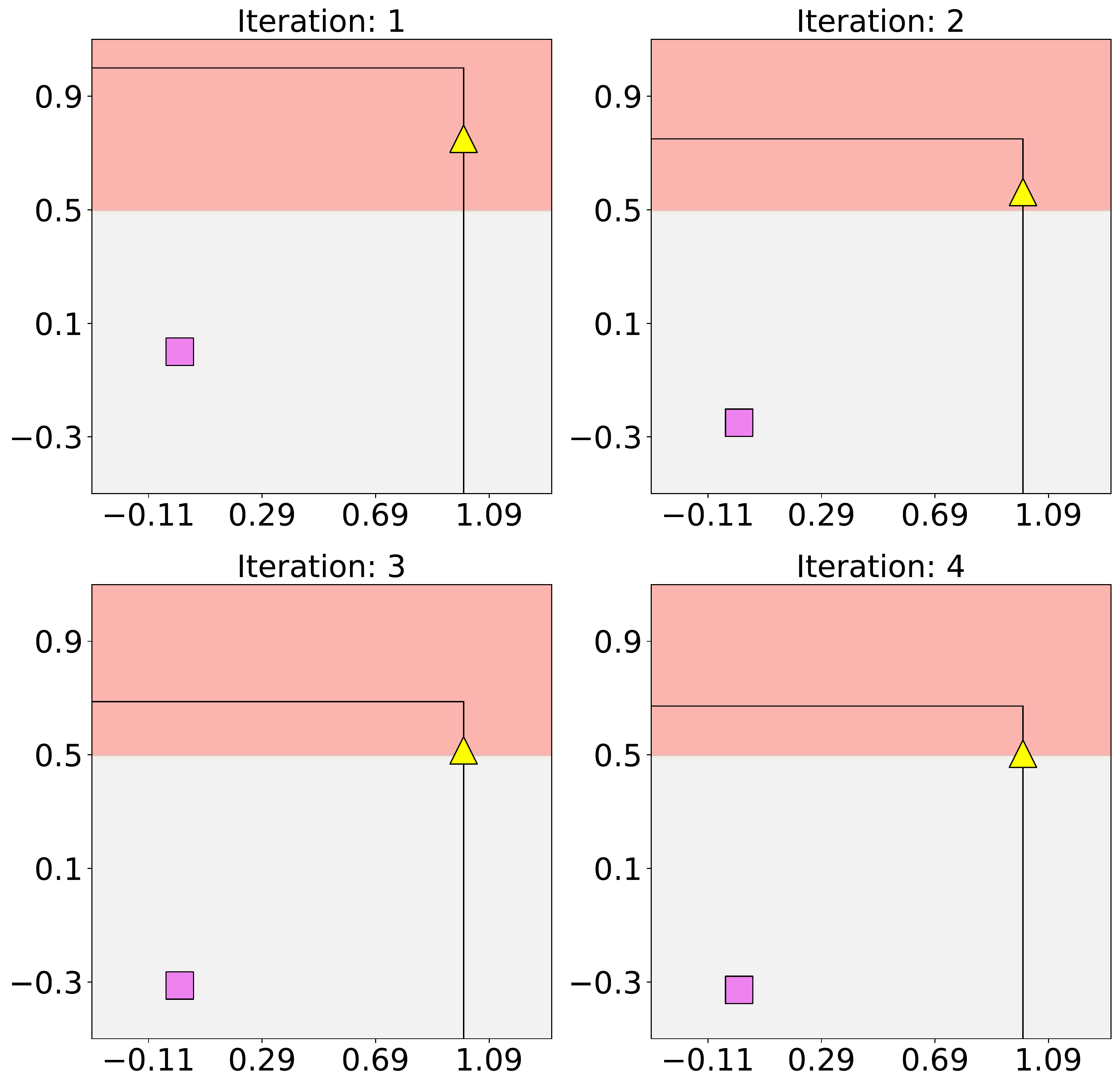}
    \caption{Illustration of the iterations of Algorithm~\ref{alg:adv_algo} for the problem in Example~\ref{ex:non-convergence}. The purple square represents the current (\ref{eqn:master_problem}) solution, while the yellow triangle represents the solution of the adversarial problem. The distance of the adversarial solutions to the decision boundary is $\left( \frac{1}{4}\right)^i$ in iteration $i$.}
    \label{fig:convergence}
\end{figure}

Indeed without Lipschitz continuity, the convergence of Algorithm \ref{alg:adv_algo} cannot be ensured. We elaborate on this necessity in the following example, for which Algorithm \ref{alg:adv_algo} does not terminate in a finite number of steps.
\begin{example}\label{ex:non-convergence}
Consider a classifier $h: \mathbb R^2 \to [0,1]$ with $h(\bm{x})=0$, if $x_2>\frac{1}{2}$ and $h(\bm{x})=1$, otherwise. The threshold is $\tau=0.5$, \textit{i.e.}, a point is classified as $1$, if $x_2\le \frac{1}{2}$ and as $-1$, otherwise. The factual instance is $\hat{\bm{z}}=(0,2)$, which is classified as $-1$. Furthermore, the uncertainty set is given as $\mathcal S=\{\bm{s}\in\mathbb R^2: \ \|\bm{s}\|_{\infty}\le 1\}$. We can warm-start Algorithm \ref{alg:adv_algo} with the \eqref{eqn:master_problem} solution $\bm{x}^1=(0,0)$. Now, assume that in iteration $i$ the optimal solution returned by \eqref{eqn:adversarial_problem} is $\bm{s}^i=(1,\frac{1}{2} + \sum_{j=1}^{i} \left(\frac{1}{4}\right)^j)$. Note that in the first iteration $\bm{s}^1=(1,\frac{3}{4})$ lies on the boundary of $\mathcal S$ and $\bm{x}^1+\bm{s}^1$ is classified as $-1$, \textit{i.e.}, it is an optimal solution of \eqref{eqn:adversarial_problem}. We are looking now for the closest point $\bm{x}^2$ to $\hat{\bm{x}}$ such that $\bm{x}^2+\bm{s}^1$ is classified as $1$, that is, it has a second component of at most $\frac{1}{2}$. This is the point $\bm{x}^2=(0,-\frac{1}{4})$ which must be the optimal solution of \eqref{eqn:master_problem}. Note that $\bm{s}^2$ is again on the boundary of $\mathcal S$ and $\bm{x}^2+\bm{s}^2 = (1,\frac{1}{2}+\frac{1}{8})$ is classified as $-1$. Hence, $\bm{s}^2$ is an optimal solution of \eqref{eqn:adversarial_problem}. We can conclude inductively that the optimal solution of \eqref{eqn:master_problem} in iteration $i$ is $\bm{x}^i=(0,-\sum_{j=1}^{i} \left(\frac{1}{4}\right)^j)$ and that $\bm{s}^i$ is an optimal solution of \eqref{eqn:adversarial_problem} in iteration $i$. Note that the latter is true, since the value of $h$ is constant in the negative region, and hence, each point in the uncertainty set is an optimal solution of \eqref{eqn:adversarial_problem}. If the latter solutions are returned by \eqref{eqn:adversarial_problem}, then the sequence of solutions $\bm{x}^i$ converges to the point $\bar{\bm{x}} = (0,-\frac{1}{3})$ which follows from the limit of the geometric series. However, $\bar{\bm{x}}$ is not a robust CE regarding $S$, since for instance, $\bar{\bm{x}} + (0,1)=(0,\frac{2}{3})$ is classified as $-1$. Consequently, Algorithm \ref{alg:adv_algo} never terminates. This example is illustrated in Figure~\ref{fig:convergence}.   
\end{example}

One difficulty is modeling the constraints of the form $h(\bm{x} + \bm{s}) \ge \tau$ for different trained classifiers. For decision trees, ensembles of decision trees, and neural networks, these constraints can be modeled by mixed-integer linear constraints as we present in the following section. Another difficulty is that $h$ is discontinuous for decision trees and ensembles of decision trees. To handle these models, we have to find Lipschitz continuous extensions of $h$ with equivalent predictions to guarantee convergence of Algorithm \ref{alg:adv_algo}.

\section{Trained Models}\label{sec:trained_models}
The main contribution of this section is modeling MP and AP for different types of classifiers $h$ by mixed-integer programming formulations. Furthermore, to prove convergence we need to assure that all studied classifiers $h$ are Lipschitz continuous, which is not the case for tree-based models. Hence we introduce a Lipschitz continuous classifier for tree-based models and derive the MP and AP for it.

We give reformulations for \eqref{eqn:master_problem} and \eqref{eqn:adversarial_problem} for decision trees, tree ensembles, and neural networks that satisfy the conditions needed in Theorem \ref{thm:cnvgthm} for convergence.

\subsection{Decision Trees}\label{subsec:decision_tree}
\begin{figure}
    \centering
    \includegraphics[scale=0.4]{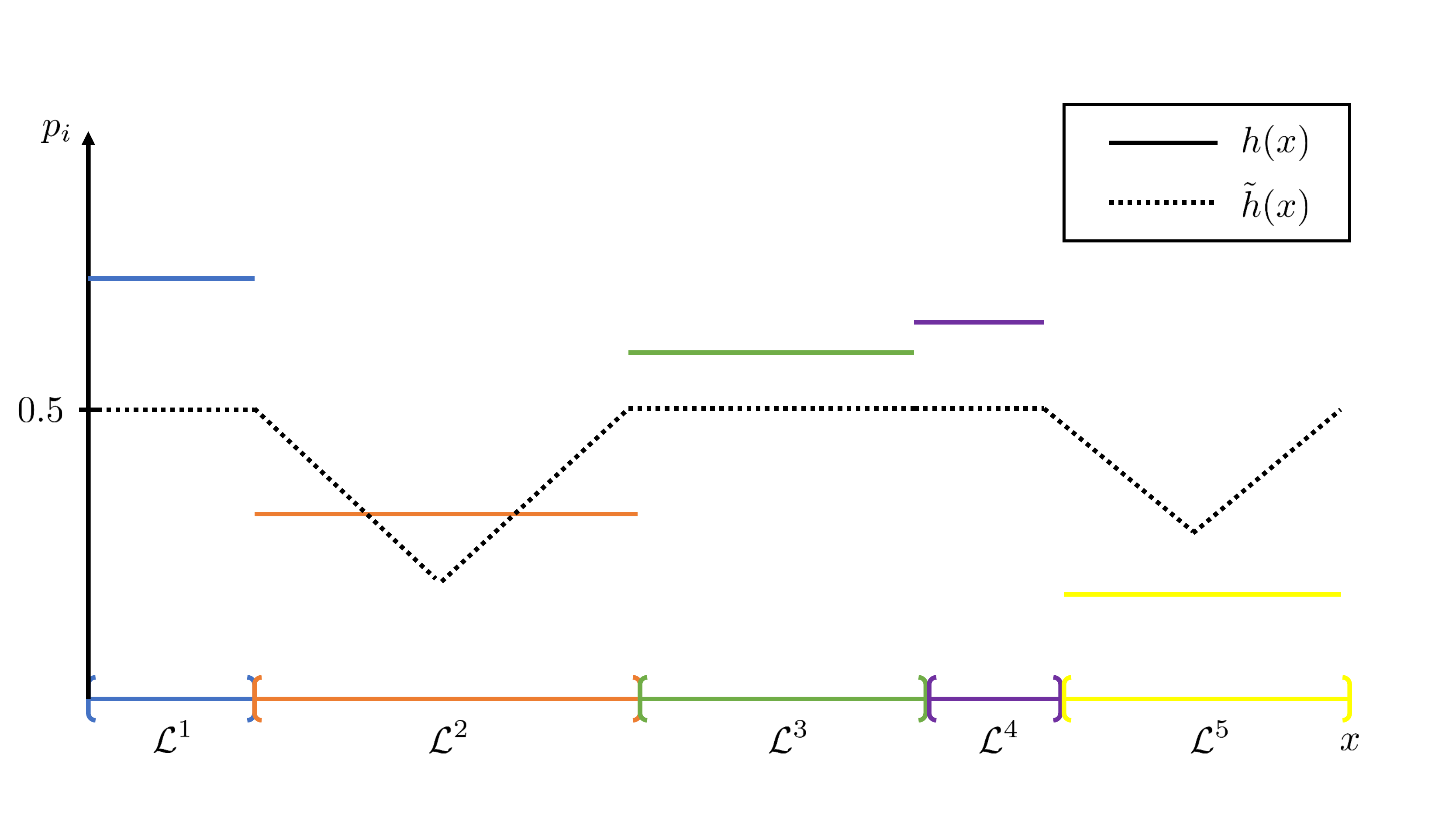}
    \caption{Example of the functions $h$ and $\tilde h$ in a one-dimensional feature space.}
    \label{fig:lipschitz_function}
\end{figure}

A decision tree (DT) partitions the data samples into distinct \textit{leaves} through a series of \textit{feature splits}. A split at node $j$ is performed by a hyperplane $\bm{\tilde a}^\top \bm{x} = \tilde b$. We assume that $\bm{\tilde a}$ can have multiple non-zero elements, in which we have the hyperplane split setting -- if there is only one non-zero element, this creates an orthogonal (single feature) split. 
Formally, each leaf $\mathcal L^i$ of a decision tree is defined by a set of (strict) inequalities
\begin{align*}
\mathcal L^i=\{ \bm{x}\in X : \bm{a}^\top \bm{x}\le  b, \ \bm{\alpha}^\top \bm{x}< \beta; (\bm{a},b)\in \mathcal N_\le^i,(\bm{\alpha},\beta)\in \mathcal N_<^i\},
\end{align*}
where $\mathcal N_\le^i$ and $\mathcal N_<^i$ contain all split parameters of the leaf for the corresponding inequality type. For ease of notation in the following, we do not distinguish between strict and non-strict inequalities and define $\mathcal N^i=N_\le^i\cup \mathcal N_<^i$.
Furthermore, it holds that $\mathbb R^n=\bigcup_{i\in L}\mathcal L^i$ where $L$ is the index set of all leaves of the tree. Each leaf $i$ is assigned a weight $p_i\in [0,1]$, which is usually determined by the fraction of training data of class $1$ inside the leaf. The classifier is a piecewise constant function $h$, where $h(\bm{x})=p_i$ if and only if $\bm{x}$ is contained in leaf $i$. Since, $h$ is a discontinuous step-function, and it is not Lipschitz continuous. To achieve convergence of our algorithm, we have to find a Lipschitz continuous function assigning the same classes to each data point as $h$. To this end we define the function 
\[
\tilde h(\bm{x})=\begin{cases} 
\tau, & \bm{x}\in\mathcal L_i, p_i\ge \tau; \\
\tau - \min\limits_{(\bm{a},b)\in \mathcal N^i} b -\bm{a}^\top \bm{x} , & \bm{x}\in\mathcal L_i, p_i< \tau.
\end{cases}
\]
We choose this function to have a constant value of $\tau$ for all leaves with $p_i\ge \tau$ while for a point $\bm{x}$ in one of the other leaves, we subtract from $\tau$ the minimum slack-value of the point over all leaf-defining constraints. Since the minimum slack on the boundary of the leaves is zero, $\tilde h$ is a continuous function and it holds $\tilde h(\bm{x})< \tau$ in the interior of the latter leaves. Note that the value of $h$ decreases if a point is farther away from the boundary of the leaf. Figure \ref{fig:lipschitz_function} illustrates this construction on a one-dimensional feature space.
Theoretically, other function classes than piecewise linear functions could be used to connect the leaf values, as long as these functions are Lipschitz continuous on each leaf region. However, this would lead to non-linear problem formulations for the adversarial problem (see below), which would increase the computational effort of our method.

Unfortunately, due to imposed continuity, the predictions on the boundaries of the leaves can be different than the original predictions of $h$. We show in the following lemma that $\tilde h$ is Lipschitz continuous and, except on the leaf boundaries, the same class is assigned to each data point as it is done by the original classifier $h$. 
\begin{lemma}\label{lem:Lipschitz}
The function $\tilde h$ is Lipschitz continuous on $\mathcal{X}$ and $\text{int}\left(\{ \bm{x}: \tilde h(\bm{x})\ge \tau\}\right) \subseteq\{\bm{x}: h(\bm{x})\ge \tau \}\subseteq \{ \bm{x}: \tilde h(\bm{x})\ge \tau\}$, where int$(\cdot)$ denotes the interior of the set.
\end{lemma}

\proof{Proof.}
We first show, that $\tilde h$ is Lipschitz continuous. To this end, let $\bm{x},\bm{y}\in \mathcal{X}$. Consider the following three cases.

\textit{Case 1:} Both points are contained in a leaf with prediction $1$, \textit{i.e.}, $\bm{x}\in \mathcal L_i$ and $\bm{y}\in \mathcal L_{i'}$ with $p_i,p_{i'}\ge \tau$. In this case we have 
\begin{align}
|\tilde h(\bm{x})-\tilde h(\bm{y})| = |\tau -\tau|=0\le \| \bm{x}-\bm{y}\|.
\end{align}
\textit{Case 2:} Point $\bm{x}$ is in a leaf with prediction $1$, and point $\bm{y}$ is in a leaf with prediction $-1$, \textit{i.e.}, $\bm{x}\in \mathcal L_i$ and $\bm{y}\in \mathcal L_{i'}$ with $p_i\ge \tau$ and $p_{i'}<\tau$. Since $\bm{x}$ is not contained in $\mathcal L_{i'}$, there must be split parameters $(a_*,b_*)$ (without loss of generality, we assume that it is related to a non-strict inequality) such that $\bm{a_*}^\top \bm{y}\le b_*$ and $\bm{a_*}^\top \bm{x}> b_*$. It holds $\tilde h(\bm{x})=\tau\ge \tilde h(\bm{y})$, and we obtain
\begin{align}
|\tilde h(\bm{x})-\tilde h(\bm{y})|&=\tau - (\tau - \min\limits_{(\bm{a},b)\in\mathcal N^{i'}} b -\bm{a}^\top \bm{y} ) \\
& = \min\limits_{(\bm{a},b)\in\mathcal N^{i'}} b -\bm{a}^\top \bm{y}  \\
&\le b_* - \bm{a_*}^\top \bm{y} \\
& < b_* - \bm{a_*}^\top \bm{y} + \bm{a_*}^\top \bm{x}-b_* \\
& = \bm{a_*}^\top (\bm{x}-\bm{y}) \\
& \le \| \bm{a_*}\| \|\bm{x}-\bm{y}\|,
\end{align}
where the first inequality follows from $(\bm{a_*},b_*)\in \mathcal N^{i'}$, the second inequality follows from $\bm{a_*}^\top \bm{x}> b_*$, and for the last inequality we apply the Cauchy-Schwarz inequality.

\textit{Case 3:} Both points are contained in a leaf with prediction $-1$, \textit{i.e.}, $\bm{x}\in \mathcal L_i$ and $\bm{y}\in \mathcal L_{i'}$ with $p_i,p_{i'}< \tau$. First assume that $i\neq i'$. Without loss of generality, we also assume that $\tilde h(\bm{x})\ge \tilde h(\bm{y})$. In this case, we have
\begin{align}
|\tilde h(\bm{x})-\tilde h(\bm{y})| \le \tau - (\tau - \min\limits_{(\bm{a},b)\in\mathcal N^{i'}} b -\bm{a}^\top \bm{y} ), 
\end{align}
which follows from $\tilde h(\bm{x})\le \tau$ for all $\bm{x}\in \mathcal{X}$. We can prove Lipschitz continuity in this case by following the same steps as in Case 2. When $i=i'$, we designate $(\bm{a_*},b_*)$ as the parameters which attain the minimum in
\begin{align}
\tau - \min\limits_{(\bm{a},b)\in\mathcal N^{i}} b -\bm{a}^\top \bm{x}.
\end{align}
Then, we have
\begin{align}
|\tilde h(\bm{x})-\tilde h(\bm{y})| &= \tau - b_* + \bm{a_*}^\top \bm{x} - (\tau - \min\limits_{(\bm{a},b)\in\mathcal N^{i'}} b -\bm{a}^\top \bm{y} ) \\
& \le   - b_* + \bm{a_*}^\top \bm{x} + b_* - \bm{a_*}^\top \bm{y} \\
& = \bm{a_*}^\top (\bm{x}-\bm{y}) \\
& \le \| \bm{a_*}\| \|\bm{x}-\bm{y}\|,
\end{align}
where we use $(\bm{a_*},b_*)\in \mathcal N^{i'}$ for the first inequality, and the Cauchy-Schwarz inequality for the last one. 

Following the three cases above, we show that $\tilde h$ is Lipschitz continuous with Lipschitz constant $L=\max_{i\in\mathcal L}\max_{(\bm{a},b)\in\mathcal N^i} \| \bm{a}\|$.

We now show the second part of the result. First, assume for $\bm{x}$ that $h(\bm{x})\ge \tau$. This implies that $\bm{x}$ is contained in a leaf $\mathcal L_i$ with $p_i\ge \tau$, and hence, $\tilde h(\bm{x})=\tau$ showing the second inclusion. For the first inclusion, let now $\bm{x}$ be a point in the interior of the set $\{ \bm{x}: \tilde h(\bm{x})\ge \tau\}$. Assume the contrary of the statement, \textit{i.e.}, it is contained in a leaf $\mathcal L_i$ with $p_i<\tau$. We can assume that the leaf is full-dimensional, since otherwise the interior is empty. Then, by definition of $\tilde h$, it must hold that
\begin{align}
\tau - \min\limits_{(\bm{a},b)\in\mathcal N^{i}} b -\bm{a}^\top \bm{x}\ge \tau.
\end{align}
That is, there is a $(\bm{a},b)\in\mathcal N^{i}$ such that $\bm{a}^\top \bm{x} = b$. Since the leaf is a full-dimensional polyhedron, there exists a $\bar \delta >0$ and a direction $\bm{v}$ such that $\bm{a}^\top (\bm{x}+\delta \bm{v}) <b$ for all $0<\delta<\bar\delta$ and all $(\bm{a},b)\in\mathcal N^{i}$. Consequently, $\tilde h(\bm{x}+\delta \bm{v})<\tau$ for all $0<\delta<\bar\delta$. This implies that $\bm{x}$ cannot be in the interior of the set  $\{ \bm{x}: \tilde h(\bm{x})\ge \tau\}$, which is a contradiction. Thus, $\bm{x}$ must be contained in a leaf with $p_i\ge \tau$ which proves the result. \hfill\Halmos\vspace{3mm}
\endproof

We can now derive the formulations for (\ref{eqn:master_problem}) and (\ref{eqn:adversarial_problem}) for our tree model. By using Lemma \ref{lem:Lipschitz}, we can use $h$ instead of $\tilde h$ to model Constraints \eqref{eqn:master_problem_constr} in (MP). Then, we can adapt the decision tree formulation proposed by \citet{maragno2022counterfactual} and reformulate Constraint \eqref{eqn:master_problem_constr} of (\ref{eqn:master_problem}) as
\begin{align}
&\bm{a}^\top (\bm{x} + \bm{s}) - M(1-l_i(\bm{s}))  \leq b, \ i \in \mathcal{L}, (\bm{a},b) \in \mathcal{N}_\le^i, \bm{s} \in \mathcal{Z}, \label{eqn:big_M_MP}\\ 
&\bm{a}^\top (\bm{x} + \bm{s}) - M(1-l_i(\bm{s}))  < b, \ i \in \mathcal{L}, (\bm{a},b) \in \mathcal{N}_<^i, \bm{s} \in \mathcal{Z}, \label{eqn:big_M_MP2}\\ 
&\sum_{i \in \mathcal{L}} l_i(\bm{s})  = 1, \quad \bm{s} \in \mathcal{Z}, \label{eqn:assign_leaf}\\
& \sum_{i \in \mathcal{L}} l_i(\bm{s}) p_i \geq \tau, \quad \bm{s} \in \mathcal{Z}, \label{eqn:outcome_constr}\\
&l_i(\bm{s})\in \{ 0,1\}, \quad i \in \mathcal{L}, \bm{s} \in \mathcal{Z},\label{eqn:general_end}
\end{align}
where $M$ is a predefined large-enough constant. The variables $l_i(\bm{s})$ are binary variables associated with the corresponding leaf $i$ and scenario $\bm{s}$, where $l_i(\bm{s})=1$, if solution $\bm{x}+\bm{s}$ is contained in leaf $i$. Constraints \eqref{eqn:assign_leaf} ensure that each scenario gets assigned to exactly one leaf. Constraints \eqref{eqn:big_M_MP} and \eqref{eqn:big_M_MP2} ensure that only if leaf $i$ is selected for scenario $\bm{s}$, \textit{i.e.}, $l_i(\bm{s})=1$, then $\bm{x}+\bm{s}$ has to fulfill the corresponding constraints of $\mathcal L_i$ while the constraints for all other leaves can be violated, which is ensured by the big-$M$ value. Note that in our computational experiments we use an $\tilde \varepsilon$-accuracy parameter to reformulate the strict inequalities as non-strict inequalities. Finally, Constraints \eqref{eqn:outcome_constr} ensure that the chosen leaf has a weight $p_i$ which is greater than or equal to the threshold $\tau$. We can remove all variables and constraints of the problem related to leaves with $p_i<\tau$ together with constraint \eqref{eqn:outcome_constr}, since only leafs which correspond to label $+1$ can be chosen to obtain a feasible solution.  

\begin{figure}[t]
\centering
\includegraphics[scale=0.3]{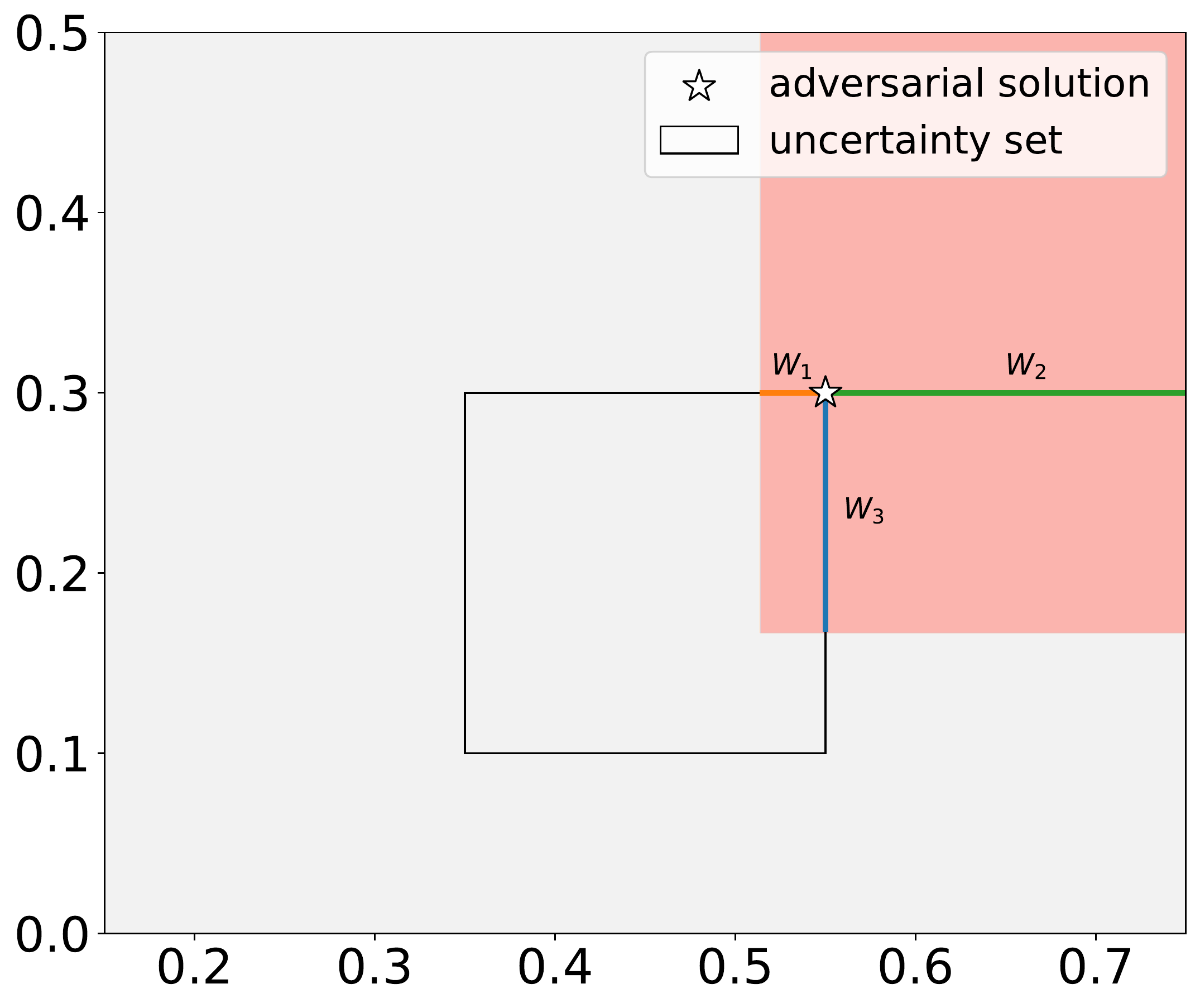}
\caption{Slack values of a solution regarding the leaf-defining constraints where the minimum slack is maximized.}
\label{fig:slacks}
\end{figure}

Using the Lipschitz continuous function $\tilde h$, (\ref{eqn:adversarial_problem}) can be reformulated as
\begin{equation}\label{eqn:APforDT}
\tau + \max_{\bm{s}\in \mathcal S} -\tilde h(\bm{x}^*+\bm{s}) .
\end{equation}
Optimizing $-\tilde h(\bm{x}^* +\bm{s})$ over $\mathcal S$ is equivalent to iterating over all leaves $\mathcal L_i$ with $p_i<\tau$ and maximizing the same function over the corresponding leaf. The problem is formulated as:
\begin{align}
\maximize &-\tau + \min\limits_{(a,b)\in \mathcal N^i}  \{b - \bm{a}^\top (\bm{x}^*+\bm{s}) \} \\
\subto &\bm{a}^\top (\bm{x}^* + \bm{s})  \leq b, \quad (\bm{a},b) \in \mathcal{N}_\le^i, \\ 
&\bm{a}^\top (\bm{x}^* + \bm{s}) < b, \quad (\bm{a},b) \in \mathcal{N}_<^i, \\
& \bm{s}\in \mathcal S
\end{align}
for each such leaf. Using a level-set transformation and substituting the latter problem in \eqref{eqn:APforDT} leads to
\begin{align}
\maximize & \alpha \label{eq:formulation_DT_slack_variables}\\
\subto  &  \alpha \le w_{(\bm{a},b)}, \quad (\bm{a},b) \in \mathcal N^i,\\
&\bm{a}^\top (\bm{x}^* + \bm{s})  + w_{(\bm{a},b)} \leq b, \quad (\bm{a},b) \in \mathcal{N}_\le^i, \\ 
&\bm{a}^\top (\bm{x}^* + \bm{s}) + w_{(\bm{a},b)} < b, \quad (\bm{a},b) \in \mathcal{N}_<^i, \\
& \bm{s}\in \mathcal S, \bm{w}\ge 0,
\end{align}
which is equivalent to maximizing the minimum slacks of the constraints corresponding to the leaves. Geometrically this means that we try to find a perturbation $\bm{s}$ such that $\bm{x}^* +\bm{s}$ is as deep as possible in one of the negative leaves; see Figure \ref{fig:slacks}. Note that the problems \eqref{eq:formulation_DT_slack_variables} are continuous optimization problems that can be solved efficiently by state-of-the-art solvers, such as, Gurobi \cite{gurobi} or CPLEX \cite{cplex2009v12}.
\vspace{3mm}

\begin{figure}[ht]
    \centering
    \begin{subfigure}{0.45\textwidth}
        \centering
        \includegraphics[width=\textwidth]{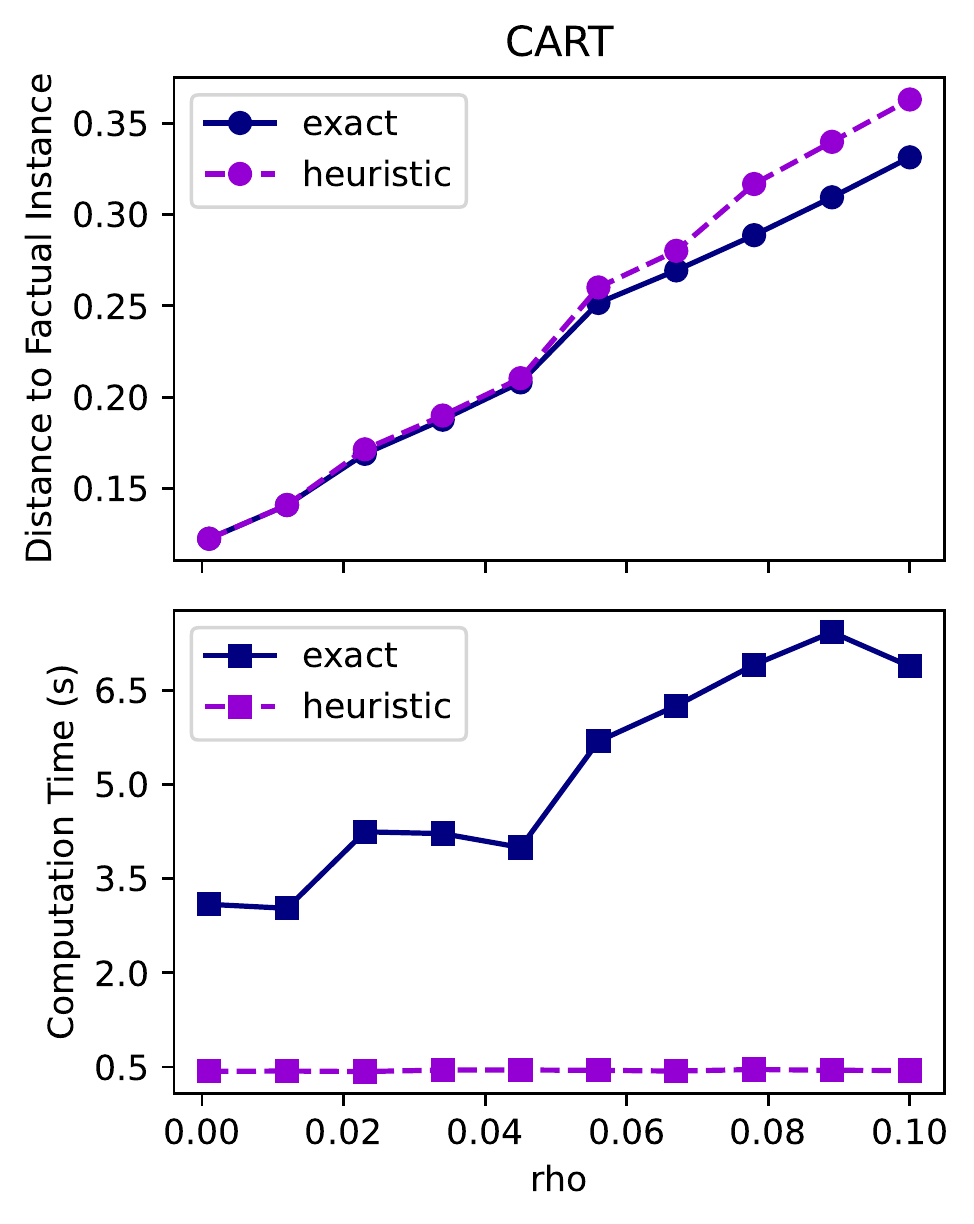}
        \caption{DT with max. depth of $5$.}
        \label{fig:exact_vs_h1}
    \end{subfigure}
    \hfill
    \begin{subfigure}{0.45\textwidth}
        \centering
        \includegraphics[width=\textwidth]{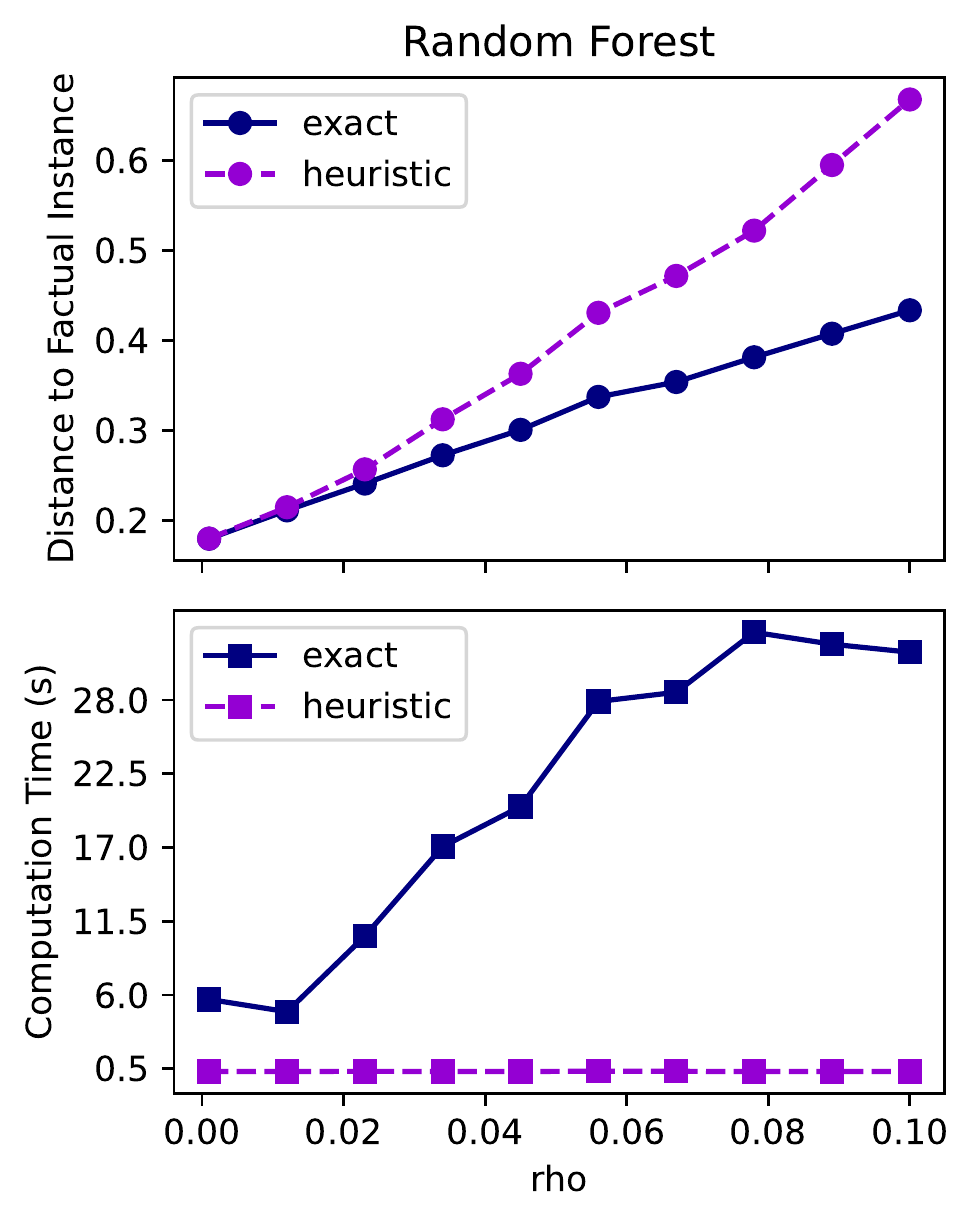}
        \caption{RF with $10$ DTs.}
        \label{fig:exact_vs_h2}
    \end{subfigure}
    \caption{Comparison of counterfactual explanations generated using the heuristic method and the adversarial algorithm (exact method) in terms of computation time and distance between the factual instance and the counterfactual explanations. The results are obtained using the Diabetes dataset and are averaged over 10 distinct factual instances.
}
    \label{fig:exact_vs_h}
\end{figure}

\begin{figure}[t]
\centering
\includegraphics[scale=0.3]{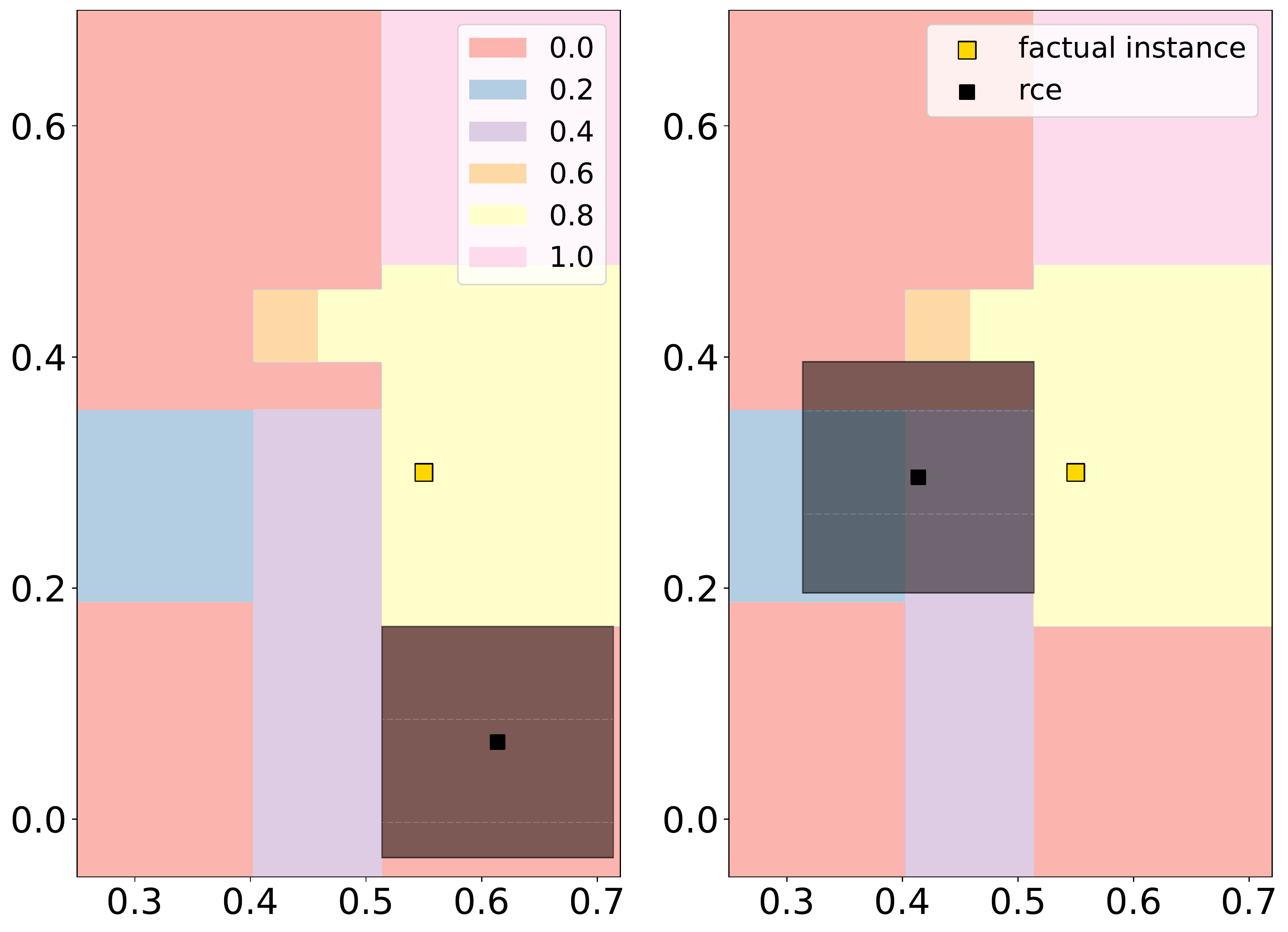}
\caption{(Left) the CE of the heuristic approach where the whole set is restricted to be contained in one leaf. (Right) an optimal CE where the uncertainty set can overlap over different leaves of a decision tree.}
\label{fig:overlappingRegions}
\end{figure}

\noindent\textbf{Heuristic Variant.}
Using Algorithm \ref{alg:adv_algo} can be computationally demanding, since it requires solving (\ref{eqn:master_problem}) and (\ref{eqn:adversarial_problem}) many times in an iterative manner. An alternative and more efficient approach can be conducted, where we try to find a CE $\bm{x}^*$ that is robust only regarding to one leaf of the tree. More precisely, this means that $\bm{x}^*+\bm{s}$ is contained in the same leaf for all $\bm{s}\in\mathcal S$. This is an approximation, since for each scenario $\bm{s}$, the point $\bm{x}^*+\bm{s}$ could be contained in a different neighboring leaf leading to better CEs; see Figure \ref{fig:overlappingRegions}. Hence, the solutions of the latter approach may be non-optimal. When restricting to one leaf, we can iterate over all possible leaves $\mathcal L_i$ with $p_i<\tau$ and solve the resulting (MP):
\begin{align}
\minimize & d(\bm{x},\bm{\hat x}) \\
\subto &\bm{a}^\top \bm{x} + \rho||\bm{a}||^*  \leq b, \quad  (\bm{a},b) \in \mathcal{N}_\le^i, \label{eqn:robust_dt}\\
    &\bm{a}^\top \bm{x} + \rho||\bm{a}||^*  < b, \quad  (\bm{a},b) \in \mathcal{N}_<^i, \label{eqn:robust_dt2} \\
    &\bm{x}\in \mathcal X,
\end{align}
and choose the solution $\bm{x}^*$ for the leaf which yields the best optimal value. Note that, in that case, we do not need binary assignment variables anymore, since we only consider one leaf for (MP). Alternatively, we can obtain the same result modelling the entire decision tree using auxiliary binary variables, one for each leaf $i$ with $p_i \geq \tau$. In Figure~\ref{fig:exact_vs_h}, we present the computation time and the distance of the calculated CE to the factual instance using both the heuristic and the (exact) adversarial algorithm. The results show that the heuristic approach outperforms the exact method in terms of speed, and its computation time remains unaffected by the robustness budget $\rho$. However, it is noteworthy that the CEs generated through the heuristic method have a larger distance to the factual instance where the difference to the optimal distance provided by our exact algorithm increases with increasing $\rho$.

\subsection{Tree Ensembles}\label{subsec:trees_ensemble}
In the case of tree ensembles like random forest (RF) and gradient boosting machines (GBM), we model the validity constraints by formulating each base learner separately. Assume we obtain $K$ base learners. Since each base learner is a decision tree, we can use the construction of the constraints \eqref{eqn:big_M_MP}-\eqref{eqn:general_end} and apply them to all base learners separately. Then, we add the constraints to the master problem, where each base learner $k$ gets a separate copy $l_i(\bm{s})(k)$ of the assignment variables and has its own set of node inequalities given by $\bm{a(k)}$ and $b(k)$. Additionally, we have to replace constraint \eqref{eqn:outcome_constr} by 
\begin{align}
    \frac{\sum_{k=1}^{K}  \sum_{i \in \mathcal{L}} l_i(\bm{s})(k) p_i(k)}{K} \ge \tau,
\end{align}
where $p_i(k)$ is the weight of leaf $i$ in base learner $k$. This constraint forces the average prediction value of the tree ensemble to be larger than or equal to $\tau$. Note that to model a majority vote, we can use $p_i(k)\in \{ 0,1\}$. Since a random forest is equivalent to a decision tree, the same methodology for (\ref{eqn:adversarial_problem}) can be used as in Section \ref{subsec:decision_tree}. Note that for classical DTs we may iterate over all leaves and solve Problem \eqref{eq:formulation_DT_slack_variables}. However, deriving the polyhedral descriptions of all leaves for an ensemble of trees is very time consuming. Instead (\ref{eqn:adversarial_problem}) can be reformulated as
\begin{align}
\maximize & \alpha \\
\subto & \alpha \le w_{(\bm{a(k)},b(k))}, \quad (\bm{a},b) \in \mathcal{N}^i(k), \ \forall k\in [K],\\
&\bm{a(k)}^\top (\bm{x}^* + \bm{s})  + w_{(\bm{a(k)},b(k))} \leq b(k), \quad (\bm{a(k)},b(k)) \in \mathcal{N}_\le^i(k), \ \forall k\in [K], \\ 
&\bm{a(k)}^\top (\bm{x}^* + \bm{s}) + w_{(\bm{a(k)},b(k))} < b(k), \quad (\bm{a(k)},b(k)) \in \mathcal{N}_<^i(k), \ \forall k\in [K], \\
& \bm{s}\in \mathcal S, \ \ \bm{w}\ge 0,
\end{align}
where $\mathcal{N}_\le^i(k), \mathcal{N}_<^i(k)$ contain the split parameters $(\bm{a(k)},b(k))$ of the nodes of tree $k$ as defined in Section \ref{subsec:decision_tree} and we use $[K]$ to denote the set of the first $K$ positive integers, that is $[K] = \{1,\dots,K\}$ .

Finally, note that since the classifier of an ensemble of trees is equivalent to a classical decision tree classifier, the convergence analysis presented in Section \ref{subsec:decision_tree} holds also for the ensemble case.

\subsection{Neural Networks}\label{subsec:neural_network}
In the case of neural networks convergence of Algorithm \ref{alg:adv_algo} is immediately guaranteed when we consider ReLU activation functions. More precisely, the evaluation function $h:\mathcal X \to [0,1]$ of a trained neural network with rectified linear unit (ReLU) activation functions is Lipschitz continuous, since it is a concatenation of Lipschitz continuous functions; see Appendix \ref{sec:appendix_proof_lipschitz_NN} for a formal proof.

Moreover, neural networks with ReLU activation functions belong to the MIP-representable class of ML models \citep{Grimstad2019, Anderson2020}, and we adopt the formulation proposed by \cite{Fischetti.2018}. The ReLU operator of a neuron in layer $l$ is given by
\begin{align}
    v_i^l = \max \left\{0, \beta_{i0}^l + \sum_{j \in N^{l-1}}  \beta_{ij}^l v_j^{l-1} \right\},\label{eqn:nnReLU}
\end{align}
where $\bm{\beta}_{i}^l$ is the coefficient vector for neuron $i$ in layer $l$, $\beta_{i0}^l$ is the bias value and $v_j^{l-1}$ is the output of neuron $j$ of layer $l-1$. Note that in our model the input of the neural network can be a data point $\bm{x}$ perturbed by a scenario $\bm{s}$, \textit{i.e.}, all variables depend on the perturbation $\bm{s}$. The input in the first layer is $\bm{v}^0(\bm{s})=\bm{x}+\bm{s}$ and the output of layer $l$ is denoted as $v_j^l(\bm{s})$. The ReLU operator (\ref{eqn:nnReLU}) can then be linearly reformulated as
\begin{align}
    v_i^l(\bm{s}) &\geq \beta_{i0}^l + \sum_{j \in \mathcal{N}^{l-1}}  \beta_{ij}^l v_j^{l-1,s}, ~~~ \bm{s}\in\mathcal{Z}, \label{eqn:relu_start} \\ 
    v_i^l(\bm{s}) &\leq \beta_{i0}^l + \sum_{j \in \mathcal{N}^{l-1}}  \beta_{ij}^l v_j^{l-1, s} + M_{LB}(1-l_i^l(\bm{s})), ~~~ \bm{s}\in\mathcal{Z}, \\
    v_i^l(\bm{s}) &\leq M_{UB}l_i^l(\bm{s}), ~ \bm{s}\in\mathcal{Z},\\
    v_i^l(\bm{s}) &\geq 0, ~~~ \bm{s}\in\mathcal{Z},\\
    l_i^l(\bm{s}) &\in \left\{ 0, 1 \right\},~~~ \bm{s}\in\mathcal{Z},\label{eqn:relu_end} 
\end{align}
where $M_{LB}$ and $M_{UB}$ are big-M values. 

The following is a complete formulation of the master problem (\ref{eqn:master_problem}) in the case of neural networks with ReLU activation functions:
\begin{align}
    \underset{{\bm{x}\in \mathcal X}}{~~\minimize} \ & d(\bm{x}, \hat{\bm{x}}) \\
    \subto &\sum_{j \in \mathcal{N}^{L}}  \beta_{j}^L v_j^{L-1}(\bm{s}) \geq \tau, ~~~ \bm{s}\in\mathcal{Z},\\
    &v_i^l(\bm{s}) \geq \beta_{i0}^l + \sum_{j \in \mathcal{N}^{l-1}}  \beta_{ij}^l v_j^{l-1}(\bm{s}), ~~~ \bm{s}\in\mathcal{Z}, ~ i \in \mathcal{N}^l, ~ \forall l\in[L], \\ 
    &v_i^l(\bm{s}) \leq \beta_{i0}^l + \sum_{j \in \mathcal{N}^{l-1}}  \beta_{ij}^l v_j^{l-1}(\bm{s}) + M_{LB}(1-l_i^l(\bm{s})), ~~~ \bm{s}\in\mathcal{Z}, ~ i \in \mathcal{N}^l, ~ \forall l\in[L], \\
   & v_i^l(\bm{s}) \leq M_{UB}l_i^l(\bm{s}), ~ \bm{s}\in\mathcal{Z}, ~ i \in \mathcal{N}^l, ~ \forall l\in[L],\\
    &v_i^{0}(\bm{s}) = x_i + s_i ,~~~ \bm{s}\in\mathcal{Z}, ~ \forall i \in [n], \\
    &v_i^l(\bm{s}) \geq 0, ~~~ \bm{s}\in\mathcal{Z}, ~ i \in \mathcal{N}^l, ~ \forall l\in[L],\\
    &l_i^l(\bm{s}) \in \left\{ 0, 1 \right\},~~~ \bm{s}\in\mathcal{Z}, ~ i \in \mathcal{N}^l, ~ \forall l\in[L],
\label{eqn:relu_app_mp}
\end{align}
where $L$ represents the number of layers with $[L] = \{1, \dots, L\}$ and  $\mathcal{N}^l$ the set of neurons in layer $l$. The first $L-1$ layers are activated by a ReLU function except for the output layer, which consists of a single node that is a linear combination of the node values in layer $L-1$. The variable $v_i^l(\bm{s})$ is the output of the activation function in node $i$, layer $l$, and scenario $\bm{s}$.

Likewise, the adversarial problem (\ref{eqn:adversarial_problem}) is formulated as
\begin{align}
    \underset{{\bm{s}\in \mathcal S}}{~~\maximize} &  \ \tau - \sum_{j \in \mathcal{N}^{L}}  \beta_{j}^L v_j^{L-1}, \\
    \subto &v_i^l \geq \beta_{i0}^l + \sum_{j \in \mathcal{N}^{l-1}}  \beta_{ij}^l v_j^{l-1}, ~~~ i \in \mathcal{N}^l, ~ \forall l\in[L], \\ 
    &v_i^l \leq \beta_{i0}^l + \sum_{j \in \mathcal{N}^{l-1}}  \beta_{ij}^l v_j^{l-1} + M_{LB}(1-l_i^l), ~~~  ~ i \in \mathcal{N}^l, ~ \forall l\in[L], \\
   & v_i^l \leq M_{UB}l_i^l, ~ i \in \mathcal{N}^l, ~ \forall l\in[L], \\
    &v_i^{0} = s_i+x_i^* ,~~~ i = 1, \dots, n, \\
    &v_i^l \geq 0, ~~~ i \in \mathcal{N}^l, ~ \forall l\in[L], \\
    &l_i^l \in \left\{ 0, 1 \right\},~~~ i \in \mathcal{N}^l, ~ \forall l\in[L],\\
    & \bm{s} \in \mathcal{S},
\end{align}\label{eqn:relu_app_ap}
where $\bm{x}^*$ is the counterfactual solution of \eqref{eqn:master_problem}.

\section{Experiments}\label{sec:experiments}
In this section, we aim to illustrate the effectiveness of our method by conducting empirical experiments on various datasets. The mixed-integer optimization formulations of the ML models used in our experiments are based on \citet{Maragno.2021}. In our experiments, we consistently set $\epsilon$ to 1e-7 and utilize a big $M$ value of 1e3 for decision trees and tree ensembles, while for neural networks, we employ a big $M$ value of 100. The experiments were conducted on a computer with an Apple M1 Pro processor and 16 GB of RAM. For reproducibility, our open-source implementation can be found at \href{https://github.com/donato-maragno/robust-CE}{our repository}\footnote{https://github.com/donato-maragno/robust-CE}. It is important to note that, to the best of our knowledge, the present work is the first approach that generates a region of CEs for a range of different models, involving non-differentiable models.  Therefore, a comparison to prior work is only possible for the case of neural networks with ReLU activation functions. In the last part of the experiments, we compare our method against the one proposed by \cite{Dominguez.2022} in terms of CE validity and robustness.
\input{tables/linf.tex}
\input{tables/l2.tex}
In the first part of the experiments, we analyze our method using three well-known datasets: \textsc{Banknote Authentication}, \textsc{Diabetes}, and \textsc{Ionosphere} \citep{Dua.2019}. Before training the ML models, we scaled each feature to be between zero and one. None of the datasets contain categorical features, which otherwise would have been considered immutable or fixed according to the user's preference. We apply our algorithm to generate robust CEs for 20 factual instances randomly selected from the dataset. We use $\ell_\infty$-norm as uncertainty set with a radius ($\rho$) of 0.01 and 0.05. The distance function adopted is the $\ell_1$-norm, which can be linearly expressed within the optimization model. For each instance, we use a time limit of 1000 seconds. Although less practical from a user's perspective, we also report the results using $\ell_2$-norm as uncertainty set in Table~\ref{tab:l2}. The datasets used in our analysis do not require any additional constraints, such as \textit{actionability}, \textit{sparsity}, or \textit{data manifold closeness}. However, it is important to note that these types of constraints can be added to our master problem when needed by using constraints like the ones proposed in \citep{maragno2022counterfactual}. The accuracy of each trained ML model is provided in Appendix \ref{app:predictive_performance}. 

\begin{figure}[ht]
    \centering
    \begin{subfigure}{0.48\textwidth}
        \centering
        \includegraphics[width=\textwidth]{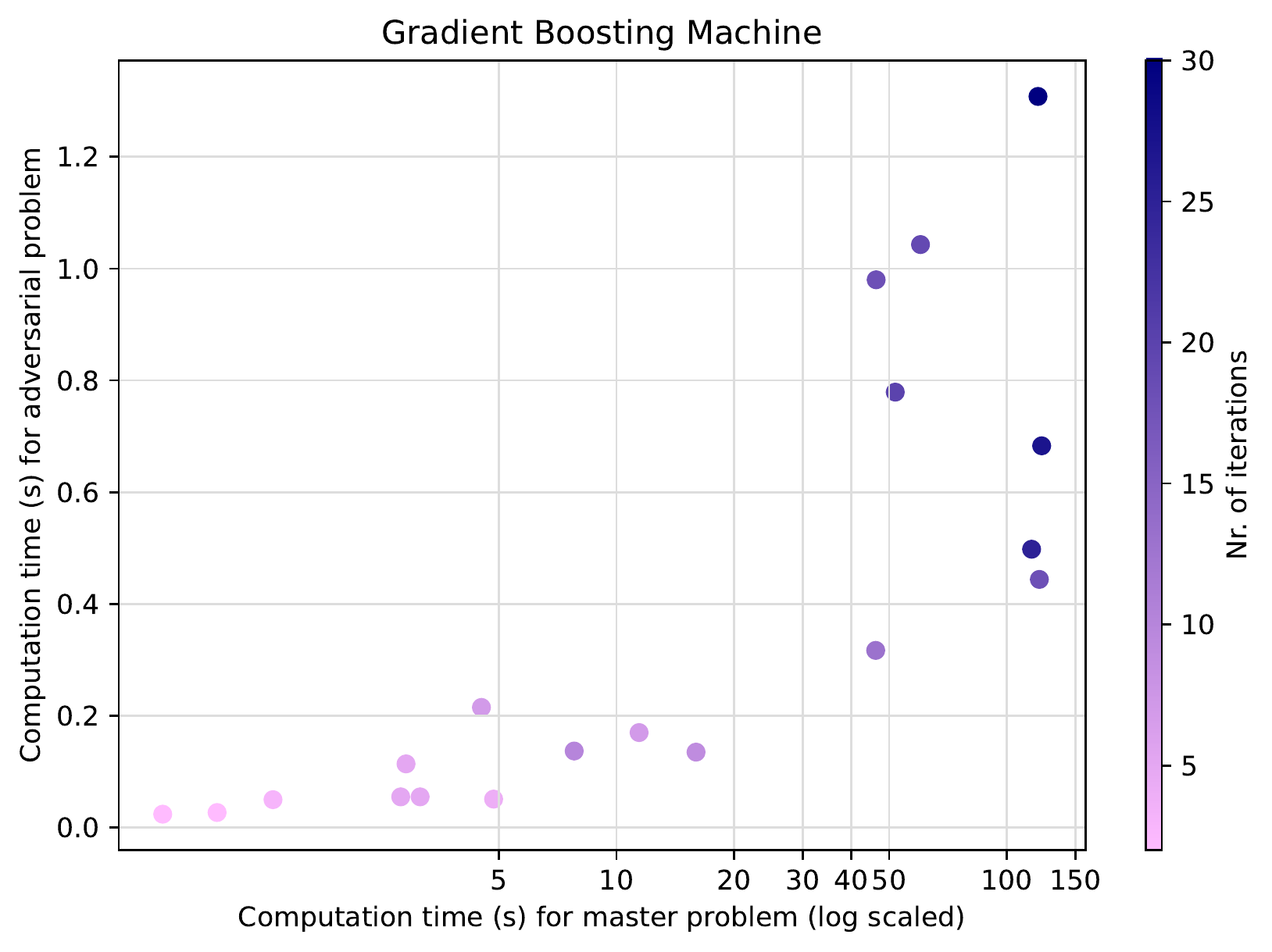}
        \caption{\small Overall computation time for 20 instances}
        \label{fig:comptimes1}
    \end{subfigure}
    \hfill
    \begin{subfigure}{0.51\textwidth}
        \centering
        \includegraphics[width=\textwidth]{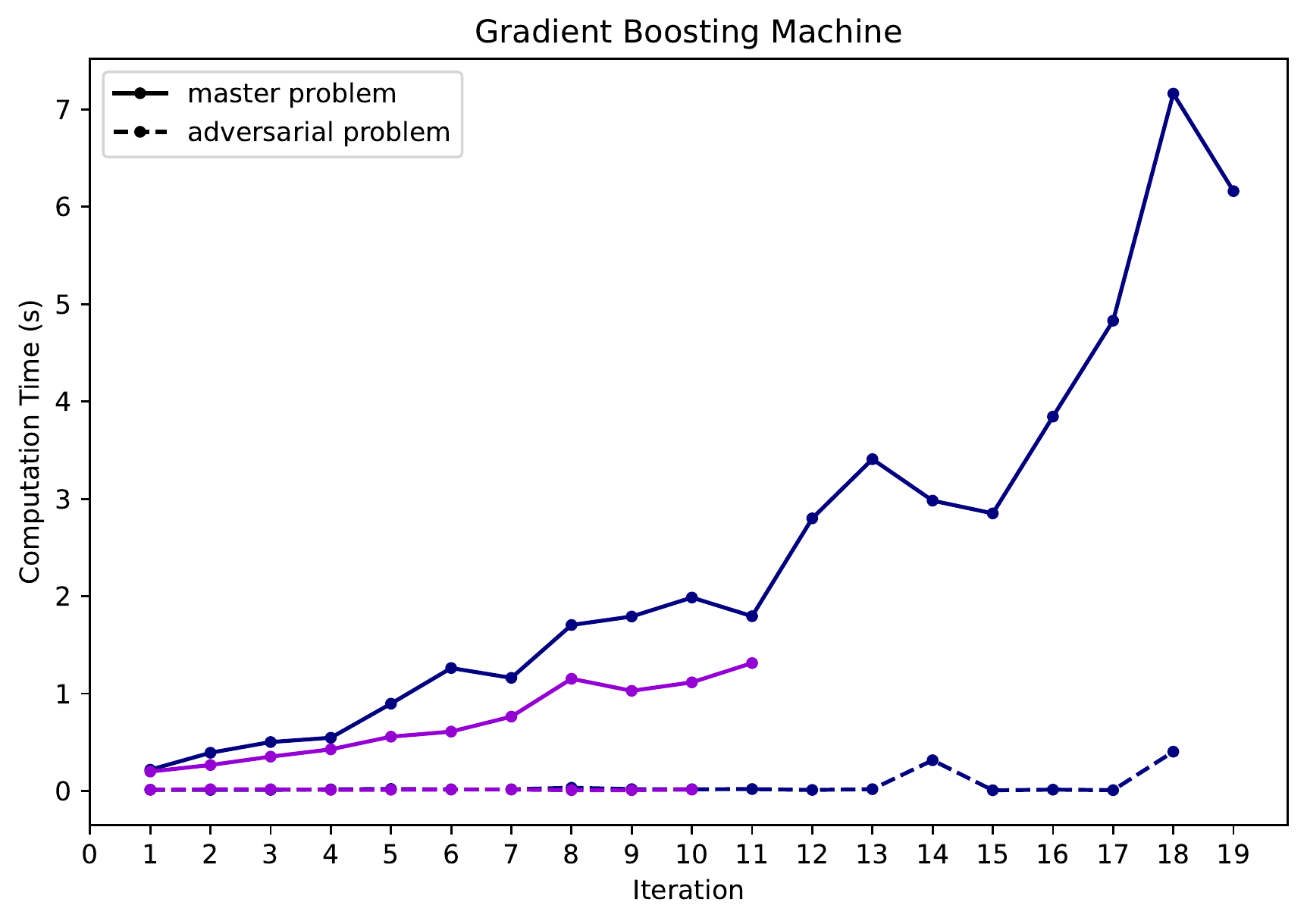}
        \caption{\small Computation time per iteration for 2 instances}
        \label{fig:comptimes2}
    \end{subfigure}
    \caption{Visual analysis of the computation time required to solve the MP and AP. The trained model is a GBM with 100 estimators. On the left, we show the computation times for MP and AP summed over all iterations, for each of 20 instances. On the right, we have visualized the computation times in each iteration of two selected instances. 
}
    \label{fig:comp_times}
\end{figure}

In Table~\ref{tab:linf}, we show (from left to right) the type of ML model, model-specific hyperparameters, and for each dataset, the average computation time (in seconds), the number of iterations performed by the algorithm, and the number of instances where the algorithm hits the time limit without providing an optimal solution. For the computation time and the number of iterations, we show the standard error values in parentheses. For the early stops, we show the (average) maximum radius of the uncertainty set, which is feasible for the generated counterfactual solutions in each iteration of the algorithm. The latter value gives a measure for the robustness of the returned solution. As for hyperparameters, we report the maximum tree depth for DT, the number of generated trees for RF and GBM, and the depth of each layer in the NN. The results indicate that the computation time and the number of iterations increase primarily due to the complexity of the ML models rather than the number of features in the datasets. 
We further inspect the computation time by looking at the time needed to solve the MP and AP at each iteration. To do so, we use the \textsc{Diabetes} dataset and train a GBM with 100 estimators. In Figure \ref{fig:comptimes1} we plot the overall computation time of the AP versus the overall computation time of the MP for 20 instances. In Figure \ref{fig:comptimes2} we inspect the time required per iteration for two of those instances.
We can see that the time required to solve the AP remains small and stable while the time required to solve the MP increases exponentially with each iteration, \textit{i.e.}, as more scenarios are added.

\begin{figure}[ht]
    \centering
    \begin{subfigure}{0.45\textwidth}
        \centering
        \includegraphics[width=\textwidth]{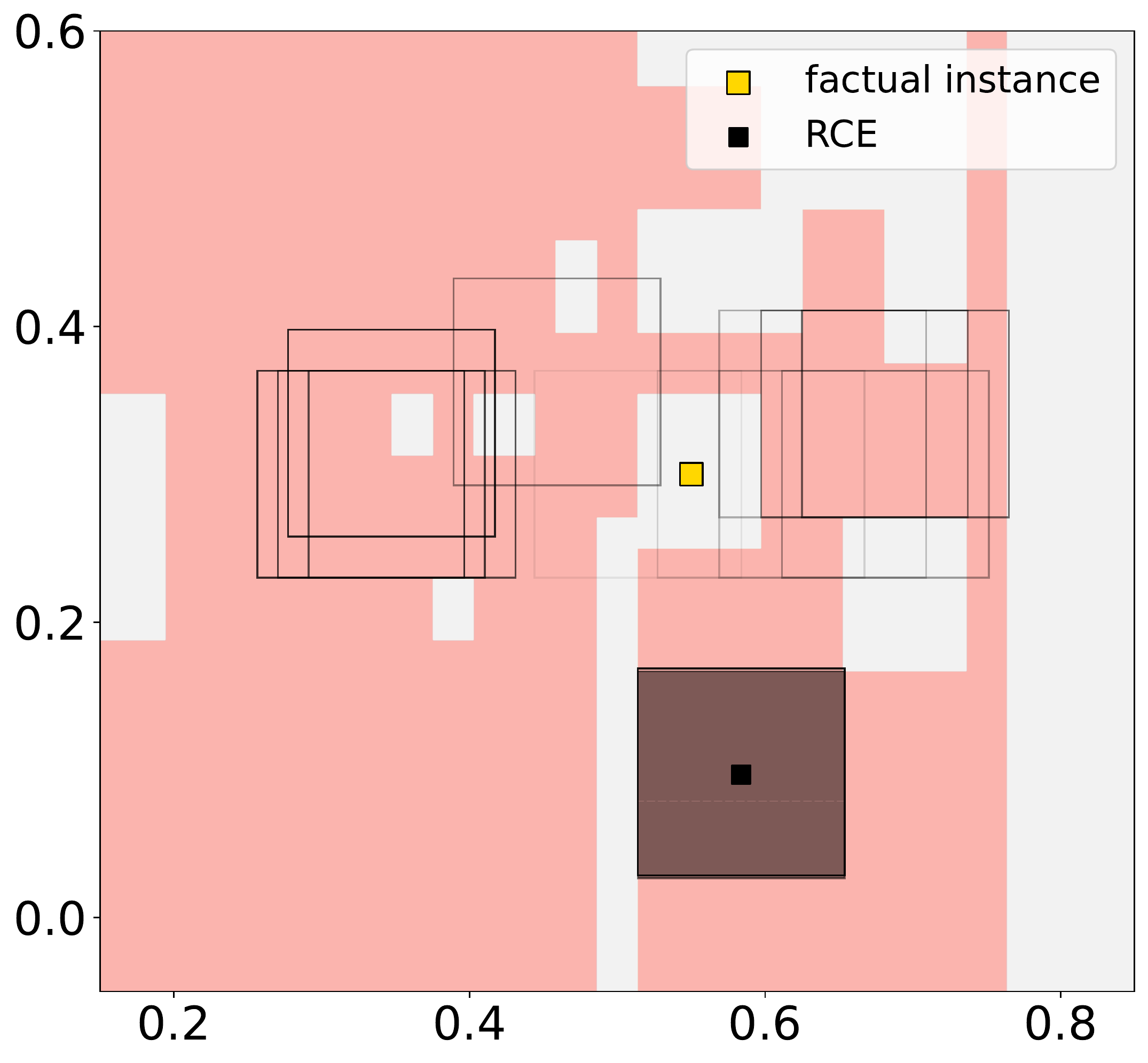}
        \caption{Decision tree with max depth = 10}
        \label{fig:overf1}
    \end{subfigure}
    \hfill
    \begin{subfigure}{0.45\textwidth}
        \centering
        \includegraphics[width=\textwidth]{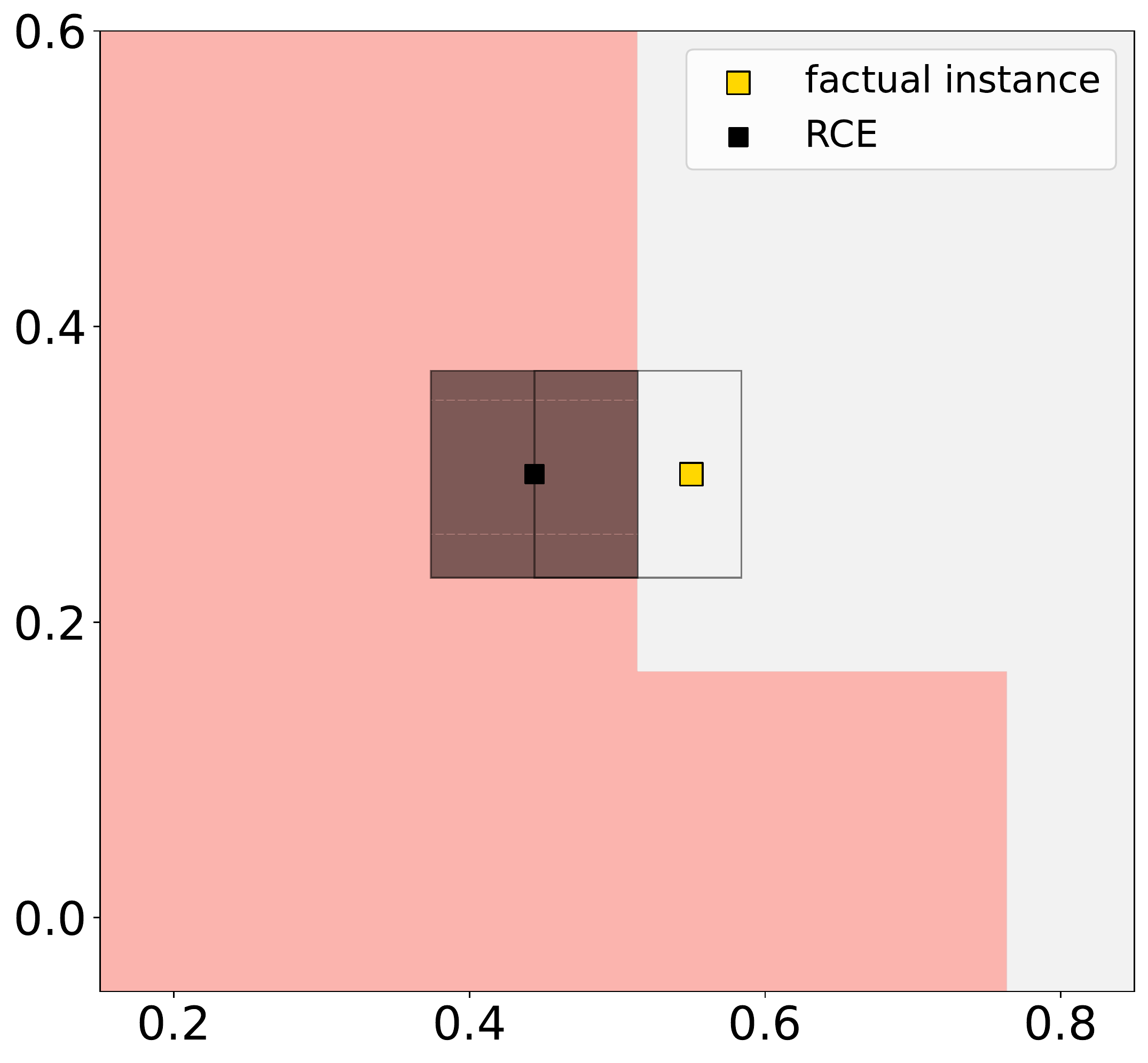}
        \caption{Decision tree with max depth = 3}
        \label{fig:overf2}
    \end{subfigure}
    \caption{Comparison of decision trees with different maximum depths. The decision tree on the right has a maximum depth of 3, while the one on the right has a maximum depth of 10 and is overfitted. The deeper decision tree leads to a higher number of iterations needed to find a robust solution, as shown by the larger number of boxes/solutions generated before converging.
}
    \label{fig:overfitting}
\end{figure}

Decision trees tend to overfit as their depth increases. In Figure~\ref{fig:overfitting}, we show iterations for a decision tree with maximum depths of (a) 10 and (b) 3.  We observe that the more complex model substantially increases the number of iterations, and hence the computation time.

\begin{figure}[hbt!]

\begin{subfigure}{.475\linewidth}
  \includegraphics[width=\linewidth]{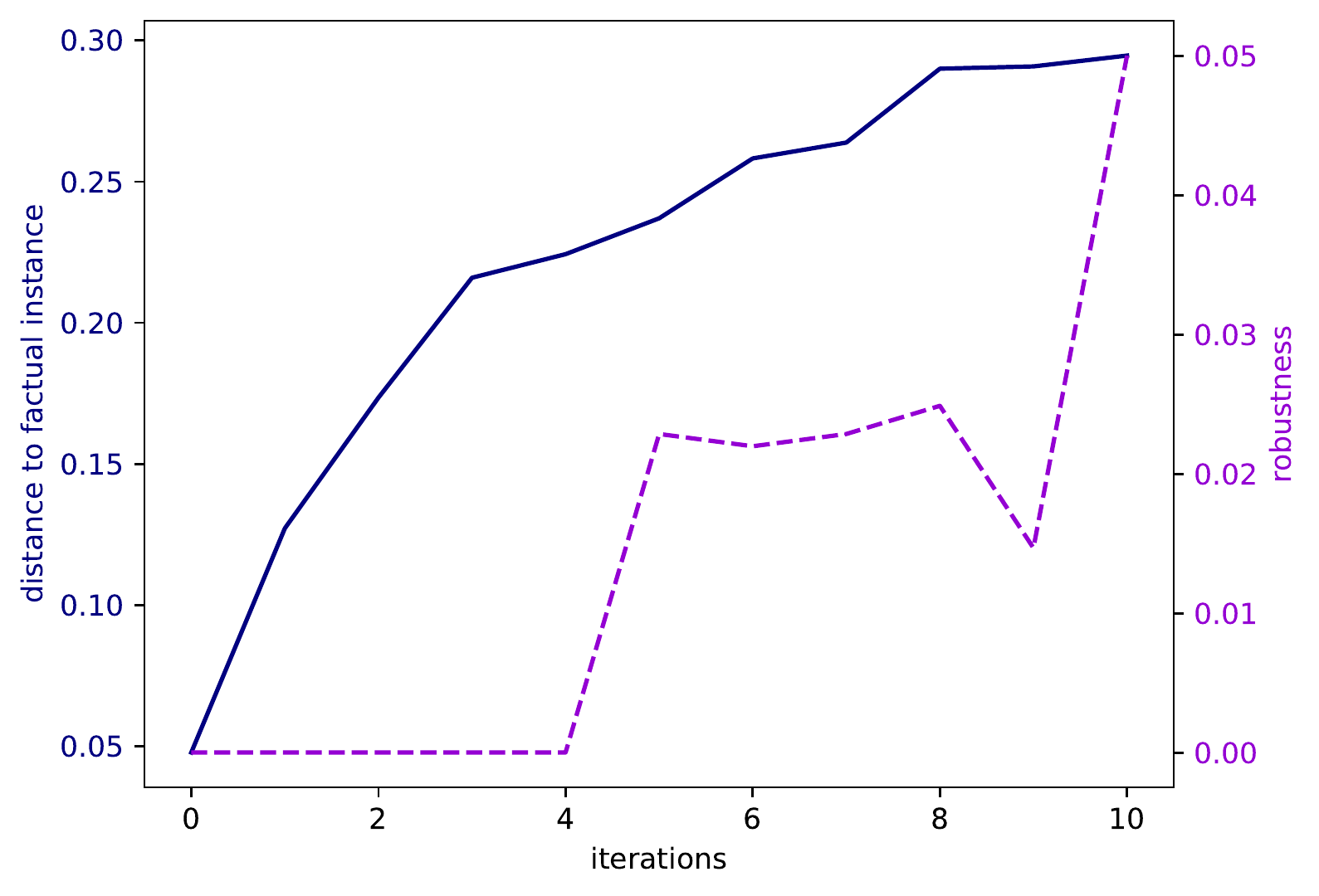}
  \caption{Decision Tree}
  \label{MLEDdet}
\end{subfigure}\hfill 
\begin{subfigure}{.475\linewidth}
  \includegraphics[width=\linewidth]{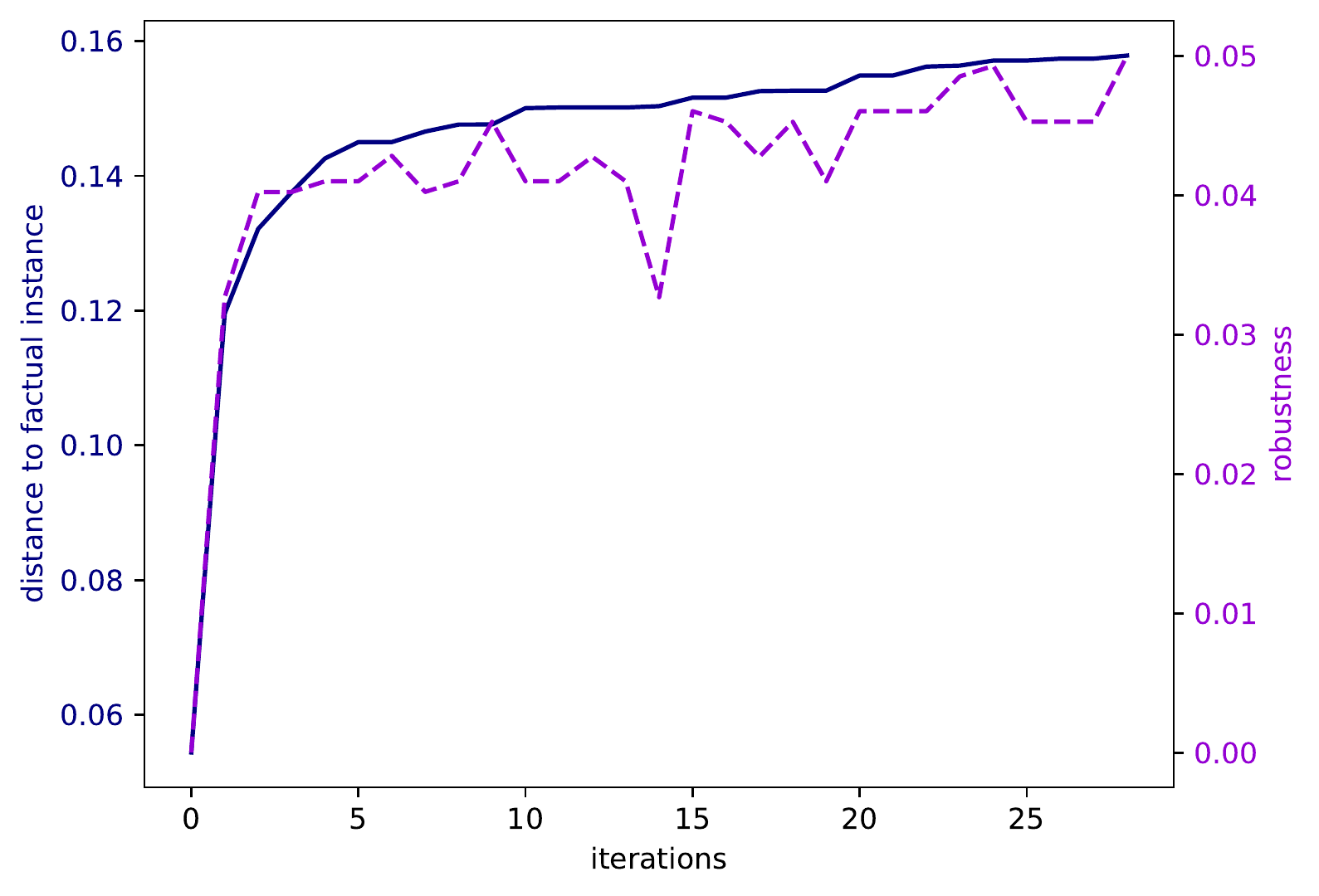}
  \caption{Random Forest}
  \label{energydetPSK}
\end{subfigure}

\medskip 
\begin{subfigure}{.475\linewidth}
  \includegraphics[width=\linewidth]{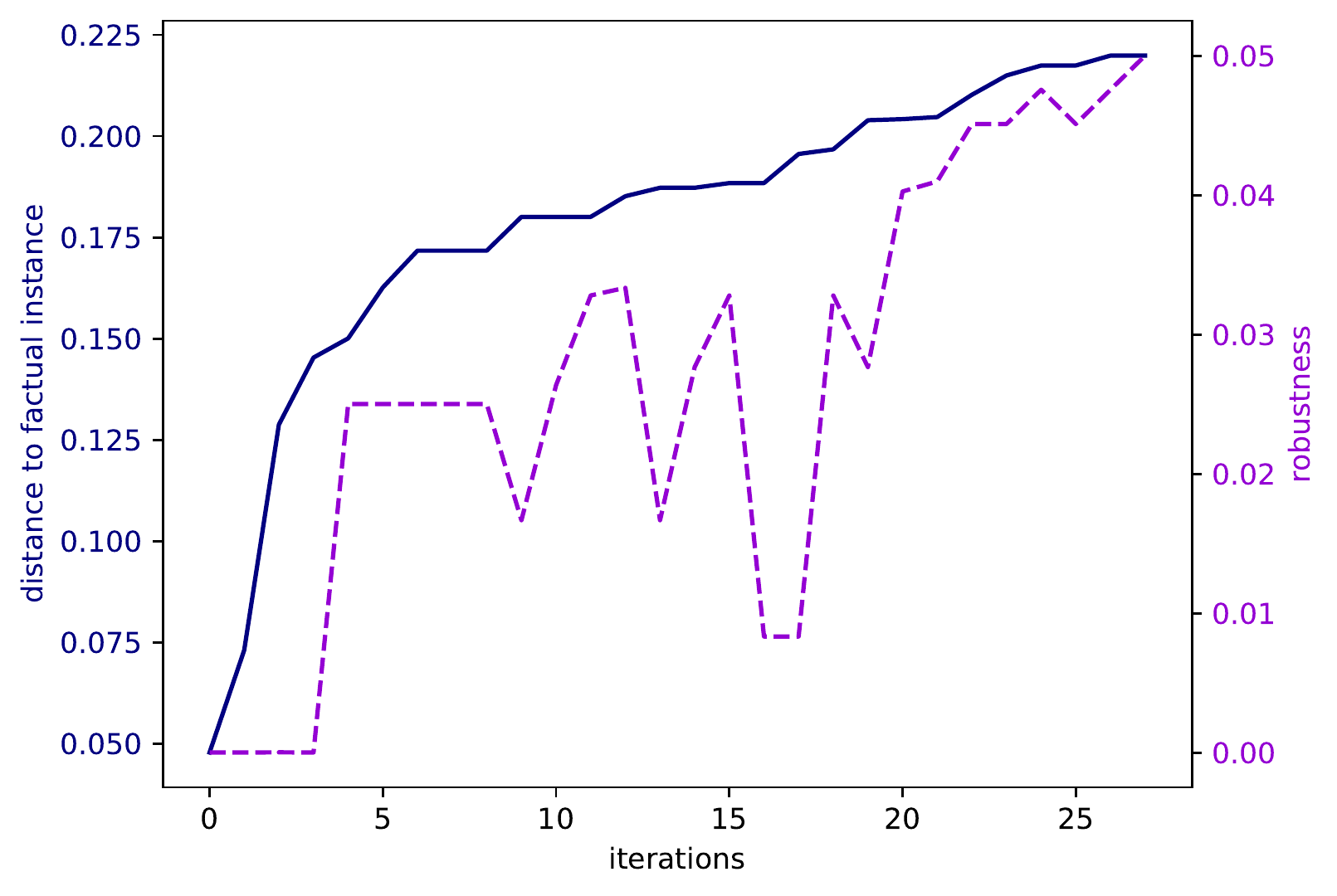}
  \caption{Gradient Boosting Machines}
  \label{velcomp}
\end{subfigure}\hfill 
\begin{subfigure}{.475\linewidth}
  \includegraphics[width=\linewidth]{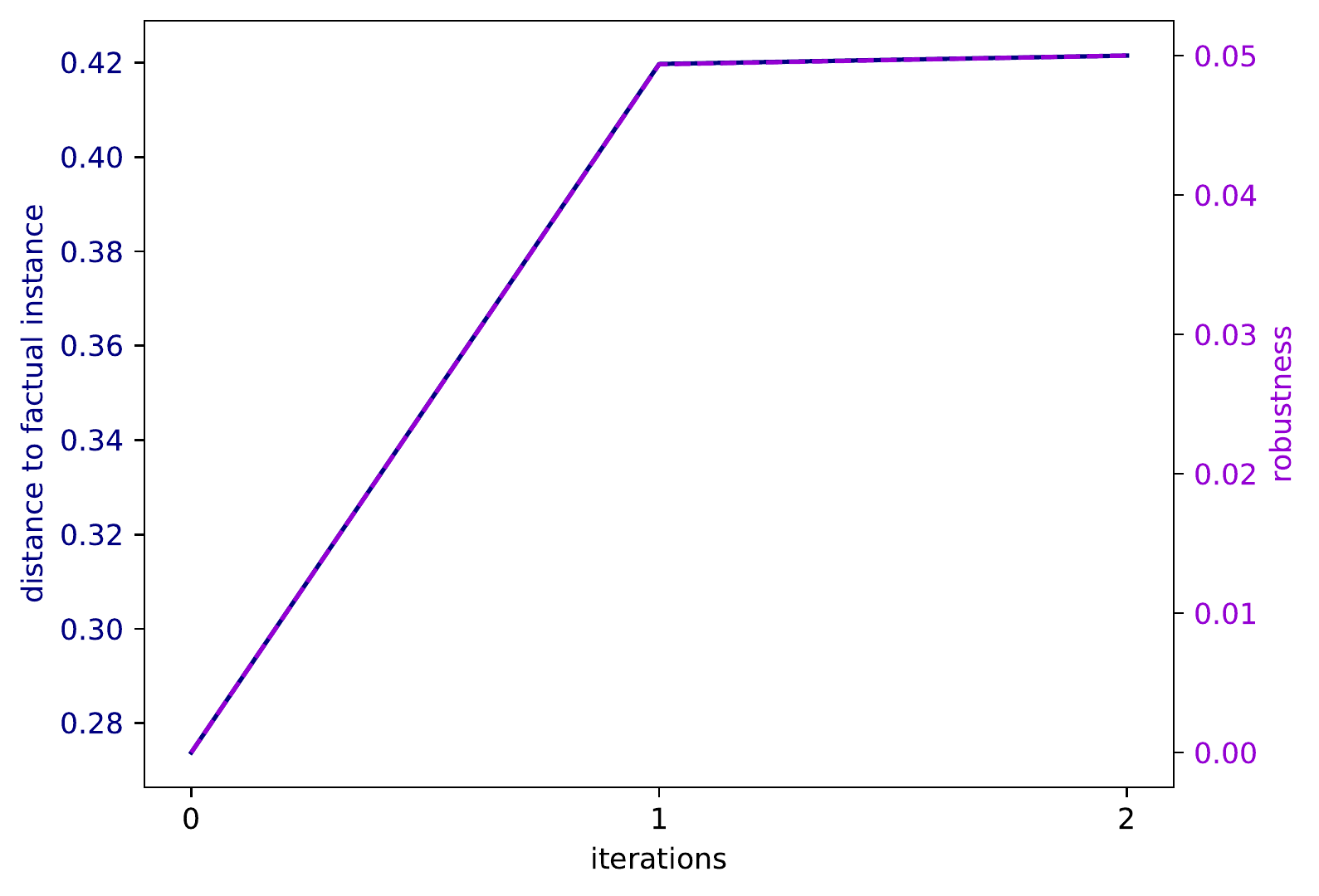}
  \caption{Neural Network}
  \label{estcomp}
\end{subfigure}

\caption{Convergence plots for DT with max depth 10 (a), RF with 50 trees (b), GBM with 50 trees (c), and NN with one hidden layer containing 100 nodes (d). The models are trained on the \textsc{Diabetes} dataset and the uncertainty set is the $\ell_\infty$-norm.}
\label{fig:convergence_plot}
\end{figure}

When the time limit is reached, we can still provide a list of solutions generated by solving the (\ref{eqn:master_problem}) at each iteration. Each one of these solutions comes with information on the distance to the actual instance and the radius of the uncertainty set for which the model predictions still belong to the desired class  for all perturbations. Therefore, the decision-makers still have the chance to select CEs from a region, albeit slightly smaller than intended. Overall the algorithm converges relatively quickly, however, our experiments show that it converges slower when using $\ell_2$-norm as uncertainty set; see Table~\ref{tab:l2}. Furthermore, as expected, for a smaller radius of $\rho = 0.01$ we reach the global optimum faster. Figure~\ref{fig:convergence_plot} shows the convergence behavior of various predictive models trained on the \textsc{Diabetes} dataset and evaluated on a specific instance. Although the distance to the factual instance (solid line) follows a monotonic increasing trend, the robustness (dashed line) exhibits both peaks and troughs. This behavior can be attributed to the objective function of \eqref{eqn:master_problem}, which seeks to minimize the distance between the CE and the factual instance. With each iteration of our algorithm, a new scenario/constraint is introduced to \eqref{eqn:master_problem}, resulting in a worse objective value (higher), but not necessarily an improvement in robustness.


\captionsetup[sub]{font=small,labelfont=small}

In the latter part of the experiment, we train a neural network with one hidden layer containing 50 nodes using the same three datasets. For the \textsc{Ionosphere} dataset, we also trained another neural network with one hidden layer containing 10 nodes. To generate robust counterfactuals, we used the $\ell_2$-norm uncertainty set and tested our algorithm against the one proposed by \cite{Dominguez.2022} on 10 instances. In Table~\ref{tab:comparison}, we report the percentage of times each algorithm was able to find a counterfactual that was at least $\rho$ distant from the decision boundary (robustness) and the percentage of times the counterfactual was valid (validity). We test $\rho$ values of 0.1 and 0.2 for each dataset. Our approach consistently generated counterfactual explanations that were optimal in terms of distance to the factual instance and valid. In the \textsc{Diabetes} dataset, our algorithm reached the time limit without finding a counterfactual with robustness of at least 0.2 in 20\% of cases. Nevertheless, the generated counterfactuals were still valid, with a $\rho$ value smaller than 0.2.
In contrast, the algorithm proposed by \cite{Dominguez.2022} often returned solutions that were not entirely robust with respect to $\rho$, particularly with the increase in $\rho$ and the complexity of the predictive model. Even more concerning, their algorithm generated invalid solutions in 10\% of the cases for the \textsc{Ionosphere} dataset. Note that our experiments only cover neural network models since the method in \citep{Dominguez.2022} cannot be used for non-differentiable predictive models.

\input{tables/comparison_dominguez}

\section{Discussion and Future Work}\label{sec:discussion}
In this paper, we propose a robust optimization approach for generating regions of CEs for tree-based models and neural networks. We have also shown theoretically that our approach converges. This result has also been supported by empirical studies on different datasets. Our experiments demonstrate that our approach is able to generate explanations efficiently on a variety of datasets and ML models. Our results indicate that the proposed method scales well with the number of features and that the main computational challenge lies in solving the master problem as it becomes larger with every iteration. Overall, our results suggest that our proposed approach is a promising method for generating robust CEs for a variety of machine learning models. 

As future work, we plan to investigate methods for speeding up the calculations of the master problem by using more efficient formulations of the predictive models, such as the one proposed by \cite{Parmentier2021} for tree-based models. Additionally, we aim to evaluate the user perception of the robust CEs generated by our approach and investigate how the choice of the uncertainty set affects the quality of the solution. Another research direction is the implementation of categorical and immutable features into the model. In the current version, categorical features are considered immutable, following user preferences. Adhering to user preferences by keeping some feature values constant, whether they are categorical or numerical, offers a practical strategy for generating sparse counterfactual explanations. This approach assumes that only a subset of the features (the ones that are not considered immutable) is susceptible to perturbations. Considering mutable categorical features would require a reformulation of the uncertainty set to account for the different feature types. This would lead to non-convex, discrete, and lower-dimensional uncertainty sets. However, this change would only affect the adversarial problem, while the master problem remains the same.

\ACKNOWLEDGMENT{%
This work was supported by the Dutch Scientific Council (NWO) grant OCENW.GROOT.2019.015, Optimization for and with Machine Learning (OPTIMAL).
}

\clearpage

%
%
%
\begin{APPENDICES}
\section{How to choose the robustness budget $\rho$}\label{app:choice_of_rho}
When determining the robustness budget $\rho$, it is essential to weigh the balance between the distance of the resulting CE to the factual instance and its robustness against small perturbations. When the underlying distribution of the CE perturbations is known, we can select $\rho$ such that a certain probabilistic guarantee is satisfied. However, in many real-world situations, we do not have access to distributional information regarding the perturbations. In this case, decision-makers can derive a Pareto front based on the CE's proximity to the factual instance and its robustness against changes. 
\subsection{Probability guarantee}
Probability guarantees regarding the invalidation of recourse depend on both the specific predictive model and the type of uncertainty set in use. For a comprehensive overview of robust reformulations with probabilistic guarantees in the context of linear models, we refer to \citep{bental2009}. In this section, we present a more generic formulation that is applicable to both linear and nonlinear models. However, it should be noted that it is a more conservative approach since it only considers the probability that the realized recourse will be in the uncertainty set while (large) part(s) of the uncertainty set may not be near the decision boundary, depending on the model structure. 

If we assume that the realized/actual perturbation $\bm{\hat{s}}$ is distributed according to $\bm{\hat{s}} \sim N(0, \sigma^2I)$, where $I$ is the identity matrix, then we can approach the probabilistic guarantee from three angles: 
\begin{enumerate}
    \item What is the probability ($\alpha$) that the realized recourse is in the defined uncertainty set $\mathcal{S} = \{\bm{s}\in\mathbb{R}^n ~~ | ~~ ||\bm{s}|| \leq \rho \}$, given $\rho$ and $\sigma$?
    \item What value of $\rho$ is required to have $\alpha$ probability that the realized recourse is in the uncertainty set $\mathcal{S} = \{\bm{s}\in\mathbb{R}^n ~~ | ~~ ||\bm{s}|| \leq \rho \}$, given $\alpha$ and $\sigma$?
    \item How large can the standard deviation of the perturbations ($\sigma$) be to have $\alpha$ probability of the realized recourse being in the uncertainty set $\mathcal{S} = \{\bm{s}\in\mathbb{R}^n ~~ | ~~ ||\bm{s}|| \leq \rho \}$, given $\alpha$ and $\rho$?
\end{enumerate}

These answers depend on the chosen norm for the uncertainty set. Note that a  prediction region with probability $\alpha$ for $\bm{\hat{s}}$ is given by $\mathcal{\tilde{S}} = \{\bm{s}\in\mathbb{R}^n ~~ | ~~ \bm{s}^T\bm{s} \leq \sigma^2\chi^2_k(\alpha)\}$ \citep{chew1966}, such that for the $l_2$ norm the questions above boil down to
\begin{align}
        \alpha = F_{\chi_k^2}\left(\frac{\rho^2}{\sigma^2}\right), \\
        \rho = \sigma\sqrt{\chi_k^2(\alpha)}, \\
        \sigma = \rho / \sqrt{\chi_k^2(\alpha)}, 
\end{align}
where $\chi_k^2(\alpha)$ and is the quantile-function for the probability $\alpha$ for the chi-square distribution with $k$ degrees of freedom, $k$ is the dimension (number of features) of $\bm{\hat{s}}$, and $F_{\chi_k^2}$ is the cumulative distribution function for the chi-square distribution with $k$ degrees of freedom.

For the $l_{\infty}$-norm, first note that $\bm{\hat{s}}$ is a vector of $k$ i.i.d. $N(0,\sigma^2)$ variables. The probability that the realized recourse is in the defined uncertainty set for this norm is then equal to the probability that all $k$ variables are between $-\rho$ and $\rho$, i.e.: $P(\bm{\hat{s}} \in \mathcal{S}) = \left[\Phi\left(\rho/\sigma\right)-\Phi\left(-\rho/\sigma\right)\right]^k = \left[2\Phi\left(\rho/\sigma\right)-1\right]^k$, where $\Phi(x)$ represents the standard normal cumulative distribution function. Equating $P(\bm{\hat{s}} \in \mathcal{S})$ to $\alpha$ then yields:
\begin{align}
        \alpha = \left[2\Phi\left(\frac{\rho}{\sigma}\right)-1\right]^k, \\
        \rho = \sigma \Phi^{-1}\left(\frac{\alpha^{1/k}+1}{2}\right), \\
        \sigma = \rho / \Phi^{-1}\left(\frac{\alpha^{1/k}+1}{2}\right), 
\end{align}
where $\Phi^{-1}(x)$ represents the inverse of the standard normal cumulative distribution function.
\subsection{Pareto front}
The Pareto front represents a set of CE solutions that cannot be improved in one objective without degrading performance in the other. In our context:
\begin{itemize}
    \item Objective 1: Proximity of the CE to the Factual Instance -- This measures the $\ell_1$-norm distance between the CE and the factual instance.
    \item Objective 2: Robustness ($\rho$) - This quantifies the robustness of the CE against perturbations.
\end{itemize}
In Figure~\ref{fig:pareto_front1}, we report the Pareto front for a decision tree with a max depth of 10 and $\ell_{\infty}$-norm uncertainty set. The outcomes are averaged across 20 different cases. When we increase the value of $\rho$, the distance to the factual instance also increases, occasionally experiencing jumps, see for example the jump in distance for values of $\rho$ around 0.06, which can be attributed to changes in the leaf nodes. Also, the computation time increases with $\rho$ given the increasing difficulty in finding a region that is large enough to contain the $\rho$-robust CE, see Figure~\ref{fig:pareto_front2}. 

\begin{figure}[ht]
    \centering
    \begin{subfigure}{0.45\textwidth}
        \centering
        \includegraphics[width=\textwidth]{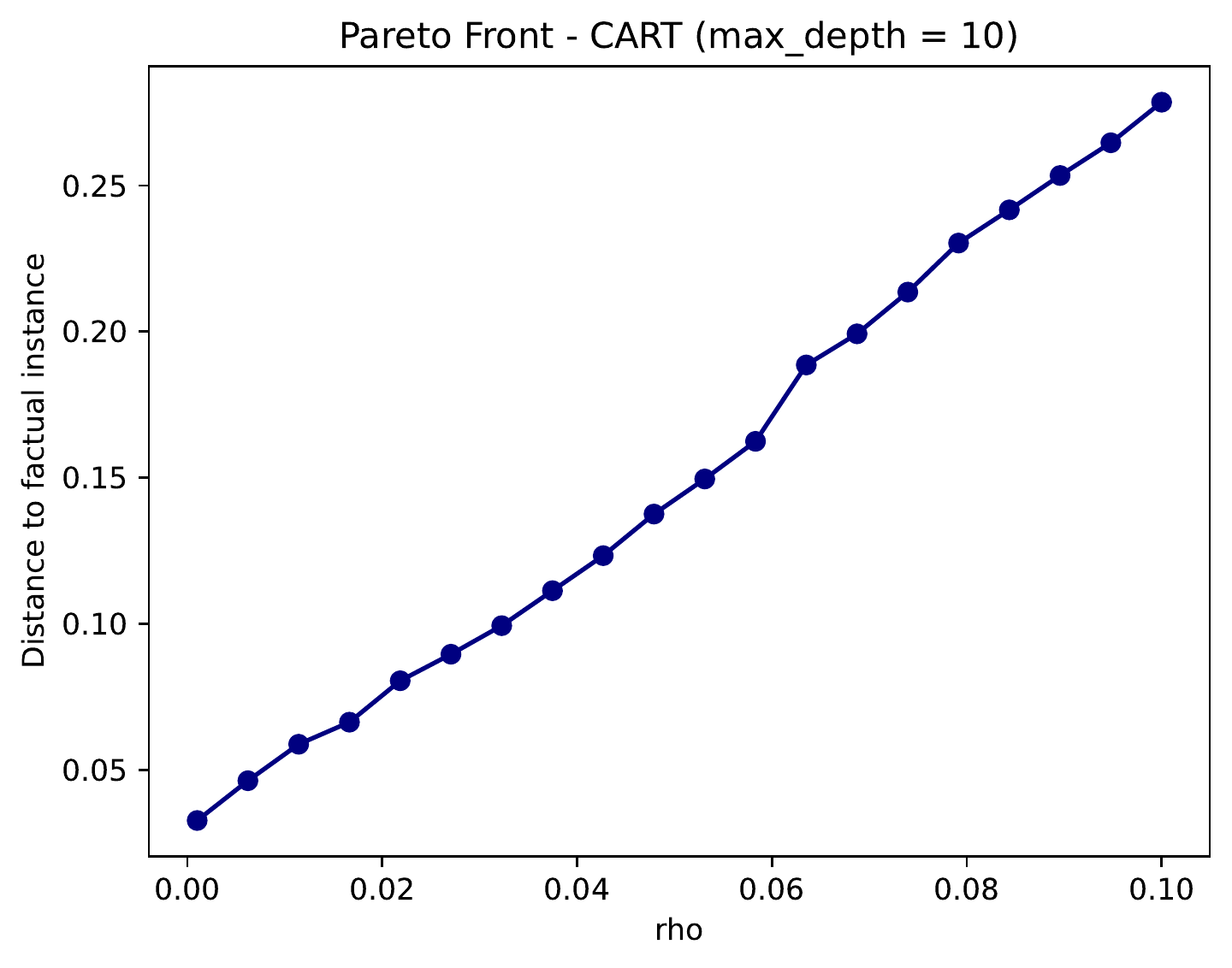}
        \caption{}
        \label{fig:pareto_front1}
    \end{subfigure}
    \hfill
    \begin{subfigure}{0.45\textwidth}
        \centering
        \includegraphics[width=\textwidth]{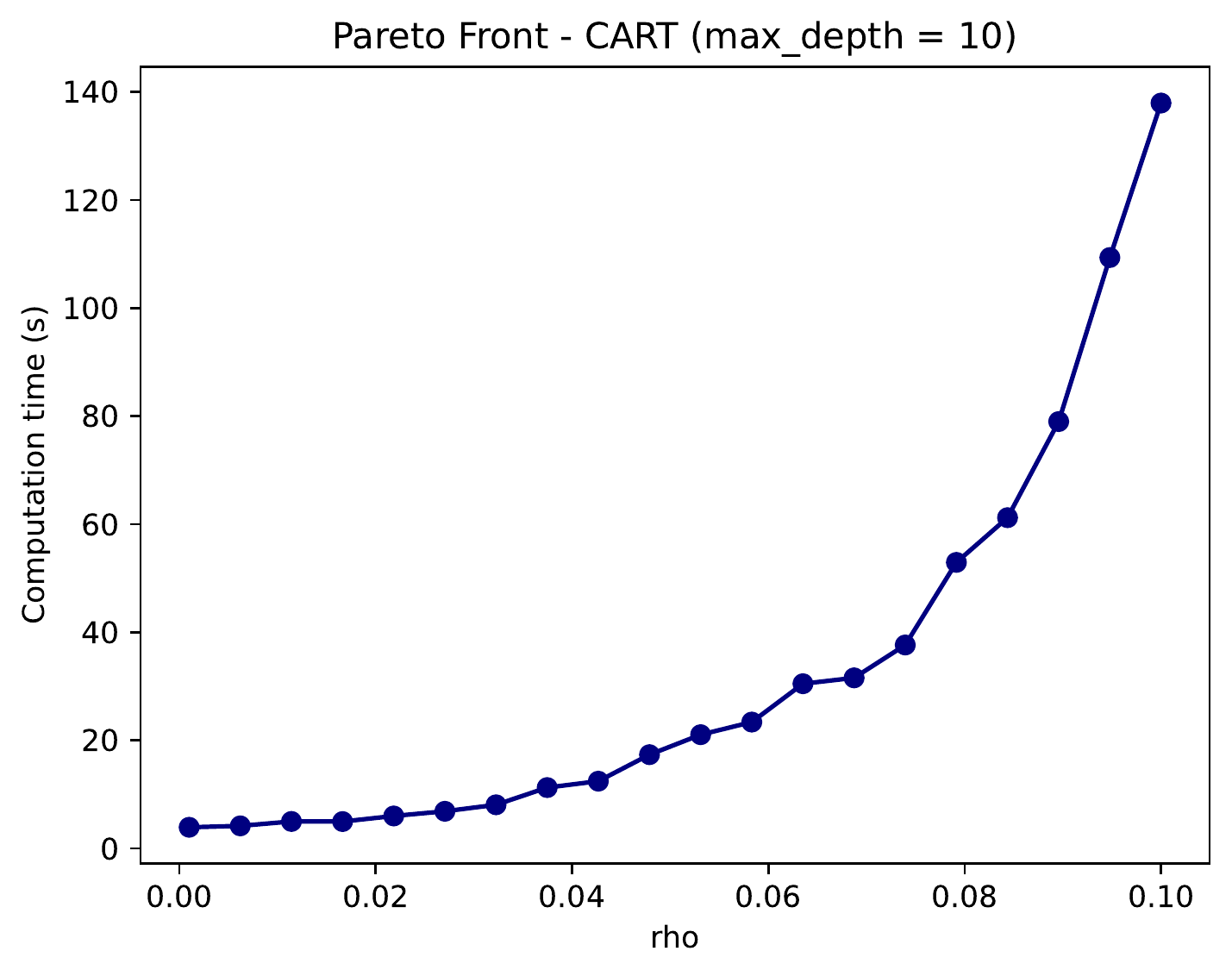}
        \caption{}
        \label{fig:pareto_front2}
    \end{subfigure}
    \caption{Pareto fronts obtained comparing (a) distance to the factual instance (y-axis) and $\rho$ (x-axis) and (b) computation time required to find a robust counterfactual (y-axis) and $\rho$ (x-axis). The results reported are averaged over 20 factual instances and obtained using a decision tree with a max-depth of 10 trained on the diabetes dataset.
}
    \label{fig:pareto_front}
\end{figure}

\section{Linear Models}\label{app:linear_model}
Although Algorithm \ref{alg:adv_algo} could still be used for the linear models, there is a well-known easier, and more efficient dual approach to solve model \eqref{eqn:RCE0}-\eqref{eqn:RCE1}. We review this approach briefly here for the completeness of our discussion.

In the case of linear models, such as logistic regression (LR) or linear support vector machines (SVM), the validity constraint~(\ref{eqn:RCE1}) can be formulated as
\begin{align}
    \bm{\beta}^\top(\bm{x} + \bm{s}) + \beta_0 \geq \tau, ~~~ \forall \bm{s} \in \mathcal{S},\label{eqn:RCE2}
\end{align}
where $\bm{\beta} \in \mathbb{R}^n$ is the coefficient vector and $\beta_0 \in \mathbb{R}$ is the intercept. Then, these constraints can be equivalently reformulated as
\[
\bm{\beta}^\top\bm{x} + \beta_0 + \min_{\bm{s}\in\mathcal S} \beta^\top \bm{s}\geq \tau .
\]
Since $\mathcal{S}$ is given in the form \eqref{eqn:RCE3}, the latter is equivalent to
\begin{align}
    \bm{\beta}^\top\bm{x} - \rho||\bm{\beta}||^* + \beta_0 \geq \tau,\label{eqn:rubust_linear}
\end{align}
where $\| \cdot \|^*$ is the dual norm of the norm used in the definition of $\mathcal S$. The constant term $\rho||\bm{\beta}||^{*}$ ensures that the constraint \eqref{eqn:RCE2} holds for all $\bm{s}\in \mathcal S$. Note that constraint~\eqref{eqn:rubust_linear} remains linear in $\bm{x}$ independently of $\mathcal S$. For more details see, \textit{e.g.}, \cite{Dominguez.2022,bertsimas2019robust,xu2008robust}.
\section{Proof of Lipschitz Continuity of Neural Networks}\label{sec:appendix_proof_lipschitz_NN}
\newcommand{\RR}{\mathbb{R}}
Suppose that we have a trained $\ell$-layer neural network constructed with ReLU activation functions. If we denote the resulting functional by $f: \RR^{n_0} \mapsto \RR^{n_\ell}$, then we can write
\begin{align}
    f(\bm{x}) = \sigma_\ell (W_{\ell} \sigma_{\ell-1}(W_{\ell-1}\sigma_{\ell-2} \dots W_2\sigma_1(W_{1}\bm{x})\dots)),
\end{align}
where $\sigma_m:\RR^{n_m} \mapsto \RR^{n_m}$, $m=1, \dots, \ell$ stands for the vectorized ReLU functions, and $W_m \in \RR^{n_m} \times \RR^{n_{m-1}}$, for $m=1, \dots, \ell$, are the weight matrices. This shows that $f$ is simply the composition of linear and component- as well as piece-wise linear functions. 

Given any two Lipschitz continuous functions $g : \RR^p \mapsto \RR^q$ and $h : \RR^q \mapsto \RR^s$ with respective Lipschitz constants $L_g$ and $L_h$, the composition $h \circ g : \RR^{p} \mapsto \RR^{p}$ is Lipschitz continuous with Lipschitz constant $L_hL_g$. This simply follows from observing for a pair of vectors $\bm{y}, \bm{z} \in \RR^p$ that 
\begin{align}
\|h \circ g(\bm{y}) - h \circ g (\bm{z})\| \leq L_h \|g(\bm{y}) - g(\bm{z})\| \leq L_hL_g \|\bm{y} - \bm{z}\|.
\end{align}
Next, for any $\bm{u}, \bm{v} \in \RR^{n_m}$, we have
\begin{align}
    \| W_m \bm{u} - W_m \bm{v} \| \leq \|W_m\|_s\|\bm{u} - \bm{v}\|.
\end{align}
where $\|\cdot \|_s$ is the spectral norm. As the Lipschitz constant for any vectorized ReLU function is one, we obtain the desired result by
\begin{align}
    \|f(\bm{\bar{x}}) - f(\bm{\tilde{x}})\| \leq \prod_{m=1}^\ell \|W_m\|_s \|\bm{\bar{x}} - \bm{\tilde{x}}\|
\end{align}
for any pair $\bm{\bar{x}}, \bm{\tilde{x}} \in \RR^{n_0}$. \hfill\Halmos\vspace{3mm}

\section{Performances of Predictive Models}\label{app:predictive_performance}

\input{tables/predictive_models}

Table~\ref{tab:ML_score} displays the accuracy score of each predictive model employed in the experiments, with their performance being reported for both the training and testing sets.

\end{APPENDICES}

\clearpage


\bibliographystyle{informs2014}
\bibliography{main}


\end{document}

%% file: tables/linf.tex

\begin{table*}[h]
\centering
\small
\resizebox{\textwidth}{!}{%
\begin{tabular}{llrrrrrrrrr}
\toprule \toprule
\textbf{} & \textbf{} & \multicolumn{3}{c}{\textbf{\textsc{BANKNOTE AUTHENTICATION}}} & \multicolumn{3}{c}{\textbf{\textsc{DIABETES}}} & \multicolumn{3}{c}{\textbf{\textsc{IONOSPHERE}}} \\ 

\textbf{} & \textbf{} & \multicolumn{3}{c}{4 features} & \multicolumn{3}{c}{8 features} & \multicolumn{3}{c}{34 features} \\ \midrule

\textbf{Model} & \textbf{Specs} & \textbf{Comp. time (s)} & \textbf{\# iterations} & \textbf{\# early stops} & \textbf{Comp. time (s)} & \textbf{\# iterations} & \textbf{\# early stops} & \textbf{Comp. time (s)} & \textbf{\# iterations} & \textbf{\# early stops} \\ \midrule

\multicolumn{11}{c}{$\rho = 0.01$} \\ \midrule

Linear & ElasticNet & 0.25 (0.01)  & -- & -- & 0.23 (0.01) & -- & -- & 0.25 (0.01)  & -- & -- \\

\rowcolor{gray!10} & max depth: 3 & 1.53 (0.04) &	 1.00 (0.00) 	& -- & 1.66 (0.07) & 	 1.20 (0.09) & 	--  & 1.66 (0.07) 	& 1.10 (0.07)  & --   \\
\rowcolor{gray!10} & max depth: 5 & 1.86 (0.09) &	 1.10 (0.07) &	 --  & 2.29 (0.11) 	& 1.20 (0.09)  & -- & 1.96 (0.08) &	 1.10 (0.07) & --    \\
\rowcolor{gray!10} \multirow{-3}{*}{DT} & max depth: 10 & 3.90 (0.88) &	 2.00 (0.58) & --  & 5.77 (0.48) & 	 1.40 (0.13) &	-- &  2.86 (0.16) 	& 1.20 (0.09) & -- 	 \\

& \# est.: 5 & 3.50 (0.36) &	 1.75 (0.22) & --  & 5.89 (1.98) &	 2.60 (0.83) &	 -- & 3.79 (0.24) &	 1.70 (0.13) &	 -- \\
& \# est.: 10 & 6.74 (1.03) &	 2.60 (0.40) &	 -- & 7.20 (0.94) &	 2.35 (0.29) & -- & 8.76 (0.89) &	 2.85 (0.33) &	 -- \\
& \# est.: 20 & 21.21 (4.61) &	 4.55 (0.99) &	 -- & 33.55 (7.48) &	 6.15 (0.79) &	--  & 22.33 (2.83) &	 4.40 (0.51) & --   \\
& \# est.: 50 & 115.79 (34.35) &	 7.80 (1.65) & --  & 110.24 (32.43) &	 6.47 (1.39) &	 3 ($\bar{\rho}$ = 0.007) & 137.26 (33.37) & 	 8.20 (1.20) & --  \\
\multirow{-5}{*}{RF$^{*}$} & \# est.: 100 & 214.38 (65.07) &	 6.44 (1.31) &	 2  ($\bar{\rho}$ = 0.009) &	274.09 (71.93) & 	 8.87 (1.49) &	 5 ($\bar{\rho}$ = 0.004) & 285.62 (95.57) &	 8.27 (2.02) &	 9 ($\bar{\rho}$ = 0.004) \\

\rowcolor{gray!10} & \# est.: 5 & 2.70 (0.22) &	 1.20 (0.14)  &	 -- & 2.76 (0.22) &	 1.85 (0.17) &	 -- & 2.37 (0.15) &	 1.60 (0.13) &	 --   \\
\rowcolor{gray!10} & \# est.: 10 & 3.20 (0.30) & 	 1.45 (0.15)  &	 -- & 2.72 (0.29) &	 1.50 (0.24) & -- & 4.35 (0.44) & 	 2.75 (0.30) & --  \\
\rowcolor{gray!10} & \# est.: 20 & 5.94 (0.50) &	 2.60 (0.23) &	 -- & 4.25 (0.45) & 	 2.15 (0.28) & -- & 9.01 (1.11) &	 3.85 (0.50) & --  \\
\rowcolor{gray!10} & \# est.: 50 &   18.38 (1.62) &	 4.05 (0.35) &	-- & 24.60 (8.21) &	 5.85 (1.41) & -- & 81.33 (28.39) &	 8.90 (1.36) & --   \\
\rowcolor{gray!10} \multirow{-5}{*}{GBM$^{**}$} & \# est.: 100 & 87.11 (26.24) &	 7.28 (0.77) 	 & 2 ($\bar{\rho}$ = 0.006) & 164.32 (42.12) &	 11.63 (2.00) &	 1 ($\bar{\rho}$ = 0.004) & 137.98 (22.74) & 	 10.33 (0.97) &	 2 ($\bar{\rho}$ = 0.007) \\

 & (10,) & 1.63 (0.04) & 	 1.00 (0.00)   &	-- & 1.55 (0.06) &	 1.00 (0.00)  & -- & 2.22 (0.19) &	 1.80 (0.21) 	 & --   \\
  & (10, 10, 10) & 2.98 (0.15) &	 1.15 (0.08)  &	--  & 2.40 (0.12) &	 1.15 (0.08) & --  & 13.01 (3.57) &	 2.30 (0.40) & --    \\
   & (50,) & 2.60 (0.14) &	 1.00 (0.00)   &	--  & 2.09 (0.12) &	 1.05 (0.05) & --  & 5.75 (0.75) &	 1.20 (0.12) &	 --  \\
 \multirow{-4}{*}{NN}   &(100,) & 3.53 (0.15) &	 1.00 (0.00) 	 &	 --  &	4.36 (0.62) &	 1.10 (0.07) & -- & 61.31 (33.11) &	 1.50 (0.22) &	 10  ($\bar{\rho}$ = 0.000) \\

\midrule

\multicolumn{11}{c}{$\rho = 0.05$} \\ \midrule

Linear & ElasticNet & 0.13 (0.00) & -- & -- & 0.15 (0.01) & -- & -- & 0.14 (0.00)  & -- & -- \\

\rowcolor{gray!10} & max depth: 3 & 1.00 (0.06) &	 1.70 (0.15) 	& -- & 1.09 (0.07) &	 1.60 (0.15) &	 --  & 1.02 (0.07) & 1.30 (0.15) &	 --   \\
\rowcolor{gray!10} & max depth: 5 & 1.18 (0.11) &	 1.80 (0.22) &	-- & 1.63 (0.13) &	 2.05 (0.22) &	 --  & 1.33 (0.10) &	 1.60 (0.18) &	 --   \\
\rowcolor{gray!10} \multirow{-3}{*}{DT} & max depth: 10 & 2.22 (0.41) &	 2.95 (0.61) & --  & 8.84 (1.49) &	 4.45 (0.59) &	 -- & 3.68 (2.14) &	 2.95 (1.49) &	-- \\

& \# est.: 5 & 2.85 (0.47) &	 3.25 (0.54) & --  &	 4.29 (0.73) &	 5.25 (0.86) &	 -- & 2.27 (0.19) &	 2.35 (0.23) &	-- \\
& \# est.: 10 & 13.51 (3.03) & 	 8.95 (1.60) &	 -- & 	 14.71 (3.53) &	 8.35 (1.24) &	-- & 7.41 (1.22) &	 5.15 (0.67) &	--  \\
& \# est.: 20 & 13.86 (3.58) &	 5.53 (0.92) &	 1 ($\bar{\rho}$ = 0.041) &	 89.00 (28.17) 	& 14.00 (2.28) &	 2 ($\bar{\rho}$ = 0.045) & 47.44 (26.35) &	 9.50 (2.52) &	--  \\
& \# est.: 50 & 101.21 (24.90) &	 11.37 (1.53) &	 1 ($\bar{\rho}$ = 0.048) & 	 303.27 (93.95) &	 16.73 (2.55) &	 9  ($\bar{\rho}$ = 0.034)  & 307.72 (72.52) &	 19.31 (3.15) &	 7 ($\bar{\rho}$ = 0.044)\\
\multirow{-5}{*}{RF$^{*}$} & \# est.: 100 & 156.28 (33.21) &	 8.70 (1.02) &	-- &	 453.67 (111.27) &	 15.43 (2.19) &	 13 ($\bar{\rho}$ = 0.032) & 156.45 (116.11) &	 5.75 (2.29) &	 16 ($\bar{\rho}$ = 0.029)\\

\rowcolor{gray!10} & \# est.: 5 & 1.57 (0.15) &	 1.75 (0.24) &	 -- &	 1.50 (0.10) &	 2.05 (0.20) &	 -- & 1.97 (0.32) &	 2.65 (0.50) &	--  \\
\rowcolor{gray!10} & \# est.: 10 & 4.84 (0.58) &	 3.55 (0.46) &	 -- &	8.87 (4.44) &	 8.55 (3.21) &	 -- & 11.76 (6.51) &	 8.85 (3.27) &	--  \\
\rowcolor{gray!10} & \# est.: 20 & 12.72 (2.22) &	 8.28 (0.85) &	 2 ($\bar{\rho}$ = 0.038) &	41.81 (17.86) &	 17.05 (4.37) &	 1 ($\bar{\rho}$ = 0.025) & 19.20 (6.32) &	 9.45 (1.80) &	--   \\
\rowcolor{gray!10} & \# est.: 50 &    73.23 (27.00) &	 13.76 (1.82) &	 3 ($\bar{\rho}$ = 0.039) &	223.87 (125.98) &	 19.86 (4.23) &	 13 ($\bar{\rho}$ = 0.027) & 139.18 (56.80) &	 16.31 (3.03) &	 7 ($\bar{\rho}$ = 0.023)  \\
\rowcolor{gray!10} \multirow{-5}{*}{GBM$^{**}$} & \# est.: 100 & 274.22 (51.40) &	 17.25 (1.57) &	 4 ($\bar{\rho}$ = 0.040) &	 -- & 	-- & 	 20 ($\bar{\rho}$ = 0.022) & 537.13 (295.04) &	 15.00 (2.08) &	 17 ($\bar{\rho}$ = 0.022)\\

 & (10,) & 0.90 (0.01) &	 1.00 (0.00) &	-- &	 1.00 (0.04) &	 1.15 (0.08) &	 -- & 2.96 (0.23) &	 3.00 (0.25) &	--  \\
  & (10, 10, 10) & 1.48 (0.03) & 	 1.00 (0.00) &	--  &	 2.02 (0.26) &	 1.65 (0.21)  &	 -- & 229.69 (104.80) &	 4.90 (0.67) &	 10  ($\bar{\rho}$ = 0.039) \\
   & (50,) & 1.39 (0.06) &	 1.00 (0.00) &	--  &	 1.75 (0.13) &	 1.35 (0.11)  &	 -- & 19.06 (6.63) 	& 2.37 (0.24) & 	 1 ($\bar{\rho}$ = 0.049) \\
 \multirow{-4}{*}{NN}   &(100,) & 1.83 (0.05) 	& 1.00 (0.00) &	 --  &	 5.88 (1.19) &	 1.80 (0.16)   &	 --  & 289.62 (125.61) &	 2.70 (0.30) &	 10 ($\bar{\rho}$ = 0.000) \\

\bottomrule \bottomrule
\multicolumn{11}{l}{\footnotesize * max depth of each decision tree equal to 3.; ** max depth of each decision tree equal to 2.}\\
\end{tabular}}
\caption{Generation of robust CEs for 20 factual instances, using $\ell_\infty$-norm as uncertainty set.}
\label{tab:linf}
\end{table*}

%% file: tables/l2.tex
\begin{table*}[h]
\centering
\small
\resizebox{\textwidth}{!}{%
\begin{tabular}{llrrrrrrrrr}
\toprule \toprule 
\textbf{} & \textbf{} & \multicolumn{3}{c}{\textbf{\textsc{BANKNOTE AUTHENTICATION}}} & \multicolumn{3}{c}{\textbf{\textsc{DIABETES}}} & \multicolumn{3}{c}{\textbf{\textsc{IONOSPHERE}}} \\ 

\textbf{} & \textbf{} & \multicolumn{3}{c}{4 features} & \multicolumn{3}{c}{8 features} & \multicolumn{3}{c}{34 features} \\ \midrule

\textbf{Model} & \textbf{Specs} & \textbf{Comp. time (s)} & \textbf{\# iterations} & \textbf{\# early stops} & \textbf{Comp. time (s)} & \textbf{\# iterations} & \textbf{\# early stops} & \textbf{Comp. time (s)} & \textbf{\# iterations} & \textbf{\# early stops} \\ \midrule

\multicolumn{11}{c}{$\rho = 0.01$} \\ \midrule

Linear & ElasticNet & 0.24 (0.01)  & -- & -- & 0.24 (0.01) & -- & -- & 0.26 (0.01)  & -- & -- \\

\rowcolor{gray!10} & max depth: 3 & 2.09 (0.11) &	 1.45 (0.11)  	& -- & 2.18 (0.13) &	 1.70 (0.15) & --   & 5.54 (0.38) &	 1.20 (0.09) &	 --   \\
\rowcolor{gray!10} & max depth: 5 & 2.64 (0.14) &	 1.53 (0.12) &	 1  ($\bar{\rho}$ = 0.000) & 4.21 (0.39) &	 2.00 (0.23) 	& --  & 10.17 (0.78) &	 1.56 (0.16) &	 4 ($\bar{\rho}$ = 0.000)   \\
\rowcolor{gray!10} \multirow{-3}{*}{DT} & max depth: 10 & 4.61 (0.88) &	 2.70 (0.58) & --  & 14.17 (2.64) & 	 2.63 (0.50) &	 1 ($\bar{\rho}$ = 0.000) &  21.75 (2.58) &	 2.06 (0.26) &	 2 ($\bar{\rho}$ = 0.000)	 \\

& \# est.: 5 & 5.20 (0.62) &	 2.35 (0.34)  & --  & 7.96 (1.65) &	 3.17 (0.55) &	 2 ($\bar{\rho}$ = 0.004) & 69.94 (10.77) &	 4.41 (0.54) &	 3 ($\bar{\rho}$ = 0.006) \\
& \# est.: 10 & 10.83 (2.68) &	 3.41 (0.78) &	 3 ($\bar{\rho}$ = 0.006) & 15.20 (3.17) &	 4.20 (0.68) & --	& 237.03 (41.12) &	 5.50 (0.70) &	 4 ($\bar{\rho}$ = 0.003) \\
& \# est.: 20 & 22.59 (7.10) &	 4.15 (1.22) &	 7 ($\bar{\rho}$ = 0.004) &	104.42 (27.35) &	 9.81 (1.78) &	 4 ($\bar{\rho}$ = 0.007) & 370.21 (41.84) &	 7.33 (0.73) &	 5 ($\bar{\rho}$ = 0.004) \\
& \# est.: 50 & 61.93 (14.29) &	 3.89 (1.22) &	 11 ($\bar{\rho}$ = 0.004) & 137.37 (39.96) &	 6.50 (1.51) &	 10 ($\bar{\rho}$ = 0.004) & 570.88 (111.94) &	 9.70 (1.61) &	 10 ($\bar{\rho}$ = 0.001) \\
\multirow{-5}{*}{RF$^{*}$} & \# est.: 100 & 103.33 (31.75) &	 3.20 (0.89) &	 10  ($\bar{\rho}$ = 0.004) & 177.47 (41.01) &	 3.80 (0.86) &	 15 ($\bar{\rho}$ = 0.002) & 531.13 (121.00) & 	 6.17 (0.98) &	 14 ($\bar{\rho}$ = 0.001) \\

\rowcolor{gray!10} & \# est.: 5 & 4.50 (0.50) &	 2.50 (0.34)  &	 -- & 4.67 (0.47) &	 2.45 (0.26) & --  & 11.47 (1.48) &	 4.42 (0.66) &	 8 ($\bar{\rho}$ = 0.002)   \\
\rowcolor{gray!10} & \# est.: 10 & 6.87 (0.88) &	 3.15 (0.42)   &	 -- & 6.11 (1.04) &	 2.70 (0.48) & -- & 42.68 (6.67) &	 6.88 (0.86) &	 3 ($\bar{\rho}$ = 0.001)  \\
\rowcolor{gray!10} & \# est.: 20 & 14.02 (1.09) &	 4.40 (0.34) &	 -- & 10.97 (1.45) &	 3.65 (0.49) & --  & 58.16 (7.01) &	 9.62 (1.00) &	 4 ($\bar{\rho}$ = 0.004)  \\
\rowcolor{gray!10} & \# est.: 50 &   34.03 (3.15) &	 5.21 (0.44) &	 1 ($\bar{\rho}$ = 0.010) & 76.04 (33.09) & 	 9.35 (2.21) & --  & 192.70 (42.52) &	 14.77 (2.47) &	 7 ($\bar{\rho}$ = 0.006) \\
\rowcolor{gray!10} \multirow{-5}{*}{GBM$^{**}$} & \# est.: 100 & 133.53 (24.40) &	 8.50 (0.98) &	 6 ($\bar{\rho}$ = 0.007) & 201.13 (77.68) &	 12.08 (2.92) &	 8 ($\bar{\rho}$ = 0.005) & 415.96 (80.09) &	 17.29 (3.36) &	 13 ($\bar{\rho}$ = 0.003) \\

 & (10,) & 1.64 (0.09) & 	 1.00 (0.00)   &	-- & 1.73 (0.05) &	 1.00 (0.00) & -- & 4.73 (0.57) &	 1.27 (0.19) &	 9 ($\bar{\rho}$ = 0.004)  \\
  & (10, 10, 10) & 2.54 (0.15) 	& 1.00 (0.00)  &	--  & 2.04 (0.10) &	 1.00 (0.00) &	 6 ($\bar{\rho}$ = 0.000)  & 282.24 (193.65) &	 1.50 (0.50) &	 18 ($\bar{\rho}$ = 0.003)  \\
   & (50,) & 2.42 (0.14) &	 1.00 (0.00)    &	--  & 2.00 (0.08) &	 1.00 (0.00) &	 1 ($\bar{\rho}$ = 0.000) & 48.13 (12.76) &	 2.40 (0.51) &	 15 ($\bar{\rho}$ = 0.008) \\
 \multirow{-4}{*}{NN}   &(100,) & 3.57 (0.14) &	 1.00 (0.00)	 &	 --  &	5.28 (0.83) &	 1.15 (0.11) & --  & 352.64 (153.96) &	 1.09 (0.09) &	 9 ($\bar{\rho}$ = 0.000)  \\

\midrule

\multicolumn{11}{c}{$\rho = 0.05$} \\ \midrule

Linear & ElasticNet & 0.15 (0.00) & -- & -- & 0.34 (0.02) & -- & -- & 0.26 (0.01)  & -- & -- \\

\rowcolor{gray!10} & max depth: 3 & 1.72 (0.17) &	 2.25 (0.19) 	& -- & 6.71 (0.65) &	 2.40 (0.24) &	 --	  &  4.40 (1.35) &	 2.60 (0.90) & --   \\
\rowcolor{gray!10} & max depth: 5 & 2.69 (0.52) &	 3.95 (0.74) &	 1 ($\bar{\rho}$ = 0.026) & 20.28 (2.88) &	 5.00 (0.68) &	 1 ($\bar{\rho}$ = 0.000) & 5.91 (1.09) &	 2.26 (0.40) & 	 1 ($\bar{\rho}$ = 0.000)   \\
\rowcolor{gray!10} \multirow{-3}{*}{DT} & max depth: 10 & 4.44 (0.80) &	 4.95 (0.93) & --  & 178.35 (34.48) & 	 9.11 (1.30) &	 1  ($\bar{\rho}$ = 0.045) &  20.91 (10.20) &	 5.05 (1.51) &	--	 \\

& \# est.: 5 & 4.49 (0.64) &	 4.40 (0.64) & --  & 117.95 (28.96) &	 9.63 (1.82) &	 1 ($\bar{\rho}$ = 0.049) & 28.79 (2.76) &	 6.35 (0.51) & -- \\
& \# est.: 10 & 12.39 (2.57) &	 6.82 (1.14) &	 3 ($\bar{\rho}$ = 0.048) & 200.39 (69.46) &	 12.92 (2.93) &	 7 ($\bar{\rho}$ = 0.043)	& 232.63 (36.46) &	 10.79 (1.36) &	 1 ($\bar{\rho}$ = 0.042) \\
& \# est.: 20 & 51.29 (18.03) &	 10.17 (2.32) &	 8 ($\bar{\rho}$ = 0.049) &	149.39 (46.57) &	 12.75 (2.78) &	 12 ($\bar{\rho}$ = 0.042) & 450.46 (68.06) &	 8.50 (1.06) &	 10 ($\bar{\rho}$ = 0.027) \\
& \# est.: 50 & 93.18 (51.21) 	& 7.78 (2.17) &	 11 ($\bar{\rho}$ = 0.048) & 560.78 (--) &	 6.00 (--) &	 19 ($\bar{\rho}$ = 0.031) & 510.69 (341.34) &	 7.50 (3.50) &	 18 ($\bar{\rho}$ = 0.021) \\
\multirow{-5}{*}{RF$^{*}$} & \# est.: 100 & 108.47 (29.76) 	& 5.25 (0.80) & 	 8 ($\bar{\rho}$ = 0.047) &	868.10 (--) &	 4.00 (--) &	 19 ($\bar{\rho}$ = 0.024) & 495.03 (268.88) &	 9.00 (2.08) &	 17 ($\bar{\rho}$ = 0.014)  \\

\rowcolor{gray!10} & \# est.: 5 & 2.67 (0.33) &	 3.55 (0.47)  &	 -- & 8.89 (1.29) &	 3.68 (0.53) &	 1 ($\bar{\rho}$ = 0.000) & 11.38 (1.83) &	 5.74 (0.68) &	 1 ($\bar{\rho}$ = 0.029)  \\
\rowcolor{gray!10} & \# est.: 10 & 5.77 (0.45) &	 5.55 (0.44)) &	 -- & 46.46 (14.33) &	 12.00 (3.12) &	 --  & 72.15 (15.31) &	 15.19 (2.62) &	 4 ($\bar{\rho}$ = 0.005) \\
\rowcolor{gray!10} & \# est.: 20 & 23.89 (3.55) &	 10.41 (1.00) 	& 3 ($\bar{\rho}$ = 0.046) & 153.43 (43.11) &	 18.76 (3.73) &	 3 ($\bar{\rho}$ = 0.036) & 156.17 (19.16) &	 16.42 (1.71) &	 8 ($\bar{\rho}$ = 0.005) \\
\rowcolor{gray!10} & \# est.: 50 &   123.75 (51.10) &	 18.14 (4.47) 	& 6 ($\bar{\rho}$ = 0.044) & 243.24 (80.72) &	 24.50 (6.31) &	 14 ($\bar{\rho}$ = 0.029)  & 894.23 (100.35) &	 32.50 (6.50) &	 18 ($\bar{\rho}$ = 0.013)   \\
\rowcolor{gray!10} \multirow{-5}{*}{GBM$^{**}$} & \# est.: 100 & 389.06 (124.83) &	 20.25 (3.32) &	 12 ($\bar{\rho}$ = 0.044) & -- (--) &	 -- (--) &	 20 ($\bar{\rho}$ = 0.020)  & -- (--) &	 -- (--) &	 20 ($\bar{\rho}$ = 0.010) \\

 & (10,) & 0.91 (0.00) &	 1.00 (0.00)  &	-- & 4.13 (0.13) &	 1.00 (0.00) &	--  & 36.34 (5.22) &	 1.68 (0.19) &	 1 ($\bar{\rho}$ = 0.000)  \\
  & (10, 10, 10) & 1.45 (0.03) &	 1.00 (0.00) &	--  & 8.02 (0.86) & 	 1.31 (0.18) &	 4 ($\bar{\rho}$ = 0.024) & 328.70 (72.59) &	 2.38 (0.35) &	 4 ($\bar{\rho}$ = 0.012)   \\
   & (50,) & 1.36 (0.03) &	 1.00 (0.00)  &	--  & 11.27 (1.10) &	 1.50 (0.15) &	 -- & 57.07 (7.28) & 	 1.22 (0.10) &	 2 ($\bar{\rho}$ = 0.000) \\
 \multirow{-4}{*}{NN}   &(100,) & 2.26 (0.04) &	 1.00 (0.00)  &	 --  &	26.82 (3.39) &	 1.30 (0.13) & --  & 111.85 (26.18) &	 1.44 (0.24) &	 11 ($\bar{\rho}$ = 0.001) \\

 \bottomrule \bottomrule
\multicolumn{11}{l}{\footnotesize * max depth of each decision tree equal to 3.; ** max depth of each decision tree equal to 2.}\\
\end{tabular}}
\caption{Generation of robust CEs for 20 factual instances, using $\ell_2$-norm as uncertainty set.}
\label{tab:l2}
\end{table*}

%% file: tables/comparison_dominguez.tex
\newcolumntype{R}[1]{>{\raggedleft\let\newline\\\arraybackslash\hspace{0pt}}m{#1}}
\begin{table}[]
\centering
\resizebox{0.6\textwidth}{!}{%
\begin{tabular}{@{}p{0.05\textwidth}R{0.15\textwidth}R{0.1\textwidth}R{0.15\textwidth}R{0.15\textwidth}}
\toprule \toprule
& \multicolumn{2}{c}{\textbf{\cite{Dominguez.2022}}}                          & \multicolumn{2}{c}{\textbf{Our algorithm}}                 \\
                           \hline
\textbf{$\rho$} &
  \textbf{robustness} &
  \textbf{validity} &
  \textbf{robustness} &
  \textbf{validity} \\
  \midrule
\multicolumn{5}{c}{\textsc{Banknote} NN(50)}                                                                         \\ 
{\textbf{0.1}} & {80\%}  & {100\%} & {100\%}  & {  100\%} \\
{  \textbf{0.2}} & {  0\%} & {  100\%} & {  100\%}  & {  100\%} \\ \hline
\multicolumn{5}{c}{ \textsc{Diabetes} NN(50)}                                                                                   \\ 
{  \textbf{0.1}} & {  60\%}  & {  100\%} & {  100\%}  & {  100\%} \\
{  \textbf{0.2}} & {  0\%} & {  100\%} & {  80\%} & {  100\%} \\ \hline
\multicolumn{5}{c}{ \textsc{Ionosphere} NN(10)}                                                                                \\ 
{  \textbf{0.1}} & {  70\%}  & {  90\%}  & {  100\%}  & {  100\%} \\
{  \textbf{0.2}} & {  40\%}  & {  90\%}  & {  100\%}  & {  100\%} \\ \hline
\multicolumn{5}{c}{ \textsc{Ionosphere} NN(50)}                                                                                \\
{  \textbf{0.1}} & {  50\%}  & {  100\%} & {  100\%}  & {  100\%} \\
{  \textbf{0.2}} & {  30\%}  & {  100\%} & {  100\%}  & {  100\%} \\ 
\bottomrule \bottomrule
\end{tabular}
}\caption{Comparison between \citep{Dominguez.2022} and our algorithm regarding the percentage of times each algorithm was able to find a counterfactual at least $\rho$ distant from the decision boundary, and the percentage of times the counterfactual explanations were valid and therefore resulting in a flip in the prediction.}\label{tab:comparison}
\end{table}

%% file: tables/predictive_models.tex
\begin{table}[H]
\centering
\resizebox{\textwidth}{!}{%
\begin{tabular}{ccccccccccccccccccc}
\toprule \toprule
\multicolumn{1}{l|}{} &
  \multicolumn{1}{c|}{} &
  \multicolumn{3}{c|}{\textbf{CART}} &
  \multicolumn{5}{c|}{\textbf{RF}} &
  \multicolumn{5}{c|}{\textbf{GBM}} &
  \multicolumn{4}{c}{\textbf{NN}} \\
\multicolumn{1}{l|}{\multirow{-2}{*}{}} &
  \multicolumn{1}{c|}{\multirow{-2}{*}{\textbf{LR}}} &
  \multicolumn{1}{c}{\textbf{3}} &
  \multicolumn{1}{c}{\textbf{5}} &
  \multicolumn{1}{c|}{\textbf{10}} &
  \multicolumn{1}{c}{\textbf{5}} &
  \multicolumn{1}{c}{\textbf{10}} &
  \multicolumn{1}{c}{\textbf{20}} &
  \multicolumn{1}{c}{\textbf{50}} &
  \multicolumn{1}{c|}{\textbf{100}} &
  \multicolumn{1}{c}{\textbf{5}} &
  \multicolumn{1}{c}{\textbf{10}} &
  \multicolumn{1}{c}{\textbf{20}} &
  \multicolumn{1}{c}{\textbf{50}} &
  \multicolumn{1}{c|}{\textbf{100}} &
  \multicolumn{1}{c}{\textbf{(10)}} &
  \multicolumn{1}{c}{\textbf{(10, 10, 10)}} &
  \multicolumn{1}{c}{\textbf{(50)}} &
  \multicolumn{1}{c}{\textbf{(100)}} \\ \hline
\multicolumn{19}{c}{\textsc{Banknote}} \\
\hline
\multicolumn{1}{l|}{\textbf{Train}} &
  \multicolumn{1}{l|}{0.97} &
  0.94 &
  0.98 &
  \multicolumn{1}{l|}{1.00} &
  0.96 &
  0.97 &
  0.96 &
  0.97 &
  \multicolumn{1}{l|}{0.97} &
  0.99 &
  0.99 &
  1.00 &
  1.00 &
  \multicolumn{1}{l|}{1.00} &
  0.98 &
  0.99 &
  1.00 &
  1.00 \\
\multicolumn{1}{l|}{\textbf{Test}} &
  \multicolumn{1}{l|}{0.96} &
  0.89 &
  1.00 &
  \multicolumn{1}{l|}{1.00} &
  0.96 &
  0.93 &
  0.93 &
  0.96 &
  \multicolumn{1}{l|}{0.93} &
  0.96 &
  1.00 &
  1.00 &
  1.00 &
  \multicolumn{1}{l|}{1.00} &
  1.00 &
  1.00 &
  1.00 &
  1.00 \\
  \hline
\multicolumn{19}{c}{\textsc{Diabetes}} \\
\hline
\multicolumn{1}{l|}{\textbf{Train}} &
  \multicolumn{1}{l|}{0.77} &
  0.77 &
  0.84 &
  \multicolumn{1}{l|}{0.97} &
  0.77 &
  0.77 &
  0.79 &
  0.80 &
  \multicolumn{1}{l|}{0.80} &
  0.80 &
  0.83 &
  0.87 &
  0.93 &
  \multicolumn{1}{l|}{0.99} &
  0.78 &
  0.79 &
  0.80 &
  0.82 \\
\multicolumn{1}{l|}{\textbf{Test}} &
  \multicolumn{1}{l|}{0.72} &
  0.77 &
  0.72 &
  \multicolumn{1}{l|}{0.69} &
  0.72 &
  0.69 &
  0.69 &
  0.67 &
  \multicolumn{1}{l|}{0.71} &
  0.59 &
  0.67 &
  0.69 &
  0.59 &
  \multicolumn{1}{l|}{0.62} &
  0.69 &
  0.64 &
  0.73 &
  0.77 \\
\hline
\multicolumn{19}{c}{\textsc{Ionosphere}} \\
\hline
\multicolumn{1}{l|}{\textbf{Train}} &
  \multicolumn{1}{l|}{0.88} &
  0.93 &
  0.97 &
  \multicolumn{1}{l|}{1.00} &
  0.93 &
  0.95 &
  0.94 &
  0.96 &
  \multicolumn{1}{l|}{0.96} &
  0.99 &
  1.00 &
  1.00 &
  1.00 &
  \multicolumn{1}{l|}{1.00} &
  0.98 &
  0.99 &
  0.99 &
  0.99 \\
\multicolumn{1}{l|}{\textbf{Test}} &
  \multicolumn{1}{l|}{0.83} &
  0.88 &
  0.83 &
  \multicolumn{1}{l|}{0.83} &
  0.88 &
  0.92 &
  0.92 &
  0.92 &
  \multicolumn{1}{l|}{0.92} &
  0.88 &
  0.88 &
  0.92 &
  0.92 &
  \multicolumn{1}{l|}{0.92} &
  0.89 &
  0.92 &
  0.88 &
  0.88 \\ \bottomrule \bottomrule
\end{tabular}
}\caption{Train and test accuracy scores of the predictive models used for the experiments.}
\label{tab:ML_score}
\end{table}